\documentclass[11pt,a4paper]{article}
\usepackage[T1]{fontenc}
\usepackage{lmodern}
\usepackage[margin=0.85in]{geometry}
\usepackage{microtype}
\usepackage{amsmath,amssymb}
\usepackage{graphicx,booktabs,tabularx,array}
\usepackage[numbers,sort&compress]{natbib}
\usepackage{xcolor,caption,placeins}
\usepackage[colorlinks=true,linkcolor=black,citecolor=blue!45!black,urlcolor=blue!45!black,breaklinks=true]{hyperref}
\newcolumntype{Y}{>{\raggedright\arraybackslash}X}
\newcolumntype{P}[1]{>{\raggedright\arraybackslash}p{#1}}
\renewcommand{\arraystretch}{1.12}
="[fonts/AstroupCompanyName.otf]" at 8pt
\DeclareRobustCommand{\astroupChinese}{{\companyChinese 向上跃升（北京）科技有限公司}}

\hypersetup{pdftitle={AI for Science in Space: Orbital Laboratories, Planetary Fieldwork, and Embodied Autonomy},pdfauthor={Zizhe Wang; Jia Deng; Xin Lu; Zehao Jia; Yi Huang; Yao Zhu; Jianing Li}}
\title{\vspace{-1.5em}\textbf{AI for Science in Space:\\ Orbital Laboratories, Planetary Fieldwork,\\ and Embodied Autonomy}}
\author{\normalsize Zizhe~Wang\textsuperscript{1}, Jia~Deng\textsuperscript{1}, Xin~Lu\textsuperscript{1}, Zehao~Jia\textsuperscript{1},\\[1pt]
\normalsize Yi~Huang\textsuperscript{2}, Yao~Zhu\textsuperscript{2}, and Jianing~Li\textsuperscript{3}\\[4pt]
{\normalsize \textsuperscript{1}Astroup Lab, Beijing, China}\\[1pt]
{\normalsize \textsuperscript{2}Harbin Institute of Technology, Beijing Research Institute}\\[1pt]
{\normalsize \textsuperscript{3}Peking University}\\[4pt]
{\small \{\href{mailto:wangzizhe@astroup.com}{wangzizhe},\,
\href{mailto:dengjia@astroup.com}{dengjia},\,
\href{mailto:luxin@astroup.com}{luxin},\,
\href{mailto:jiazehao@astroup.com}{jiazehao}\}@astroup.com}\\[1pt]
{\small \href{mailto:24B903024@stu.hit.edu.cn}{24B903024@stu.hit.edu.cn}\quad
\href{mailto:ee_zhuy@zju.edu.cn}{ee\_zhuy@zju.edu.cn}\quad
\href{mailto:lijianing@pku.edu.cn}{lijianing@pku.edu.cn}}}
\date{}
\newcommand{\ReviewRecordCount}{150}
\newcommand{\ReviewCoreAICount}{35}
\newcommand{\ReviewSupportingCount}{58}
\newcommand{\ReviewTransferCount}{33}
\newcommand{\ReviewBackgroundCount}{24}

\begin{document}
\maketitle
\footnotetext[1]{Astroup Lab: \astroupChinese.\quad\href{https://astroup.com}{astroup.com}}
\vspace{-2.3em}
\begin{abstract}
Artificial intelligence for science (AI4S) is extending from interpretation of existing data to the design and execution of experiments. Space offers distinctive scientific opportunities while restricting physical support. Microgravity and extraterrestrial environments expose phenomena that terrestrial experiments cannot fully reproduce; finite samples, constrained resources, and limited opportunities for intervention complicate sustained investigation. This critical review examines orbital laboratories and planetary in situ science across biology, materials, fluid and thermal physics, and adaptive exploration. We distinguish learning from space-derived data, selecting scientific actions, executing them, and preserving experimental validity. Terrestrial self-driving laboratories provide complementary evidence on optimization, dedicated automation, robotic manipulation, and the support needed for unattended operation. The synthesis connects scientific gains already demonstrated by bounded AI systems to gaps in observation, executable actions, and continuity of valid evidence. Space robots can expand the attainable scope of autonomous science when necessary interventions exceed validated instrument capabilities and timely human support is unavailable. Their value must be established against dedicated mechanisms, redundancy, and human-supported operation. We develop a framework connecting sample history, scientific decisions, physical execution, and claim-specific validity. A worked orbital design, a planetary extension, and matched evaluation schemes separate the contributions of sensing, decision policies, and physical actions. The resulting agenda measures autonomy through valid scientific return, task coverage, intervention costs, and evidence integrity across complete campaigns.
\end{abstract}
\section{Introduction: from models of science to scientific action}
\label{sec:intro}
Artificial intelligence for science (AI4S) uses computational methods for learning, inference, search, and decision-making to advance scientific understanding. Its contributions include extracting observables from complex data, constructing predictive models, proposing candidate mechanisms or materials, and selecting informative experiments \citep{bg_wang2023}. These functions occupy different positions in the scientific process. A model that predicts a molecular property can be scientifically valuable without controlling an instrument; a system that chooses an experiment must additionally connect its prediction to a feasible intervention and an interpretable outcome. The transition from analyzing evidence to actively producing it is therefore an integration problem involving scientific reasoning, experimental design, and physical execution.

Space provides a distinctive setting for this transition. Orbital laboratories make altered gravity and prolonged exposure accessible as experimental conditions. Planetary missions investigate materials and environments whose context cannot be reproduced completely on Earth. These opportunities coexist with restricted experimental resources, limited access to equipment, and variable opportunities for human intervention. The space biological and physical sciences decadal survey emphasizes both the scientific questions enabled by these environments and the infrastructure needed to investigate them \citep{bg_nasem2023}. AI4S can help make individual experiments more informative, but its longer-term value depends on whether a scientific program can continue producing trustworthy evidence between opportunities for external support.

Robot scientists and self-driving laboratories have developed experimental feedback on Earth \citep{grd_king2009,grd_tom2024}. Spacecraft planning and execution research has addressed resource constraints and changing conditions for decades \citep{rob_remoteagent1998,rob_aspen2000,rob_casper2000}. Mars instruments have demonstrated bounded scientific choices onboard \citep{pln_francis2017,pln_lawson2025}. Orbital biology has established substantial dedicated automation and an explicit agenda for AI-supported deep-space laboratories \citep{bg_sanders2023}. These traditions connect scientific inference, experimental feedback, and resource-aware execution across terrestrial and space settings.

\subsection{Scope and organizing distinctions}
This review focuses on \emph{orbital experimentation and planetary in situ science}. Orbital coverage includes biology, materials and crystal growth, fluid and thermal physics, combustion, soft matter, and a bounded cold-atom comparison. Planetary coverage includes target selection, adaptive spectroscopy, sample acquisition and handling, field surveys, and astrobiology. Sounding-rocket, parabolic-flight, terrestrial analogue, and laboratory studies are included where they clarify the experimental lineage or the transfer problem, with their settings explicitly identified. Astronomy and remote sensing are adjacent AI4S domains; they are not the main evidence base here. Generic spacecraft control, manufacturing, and servicing enter only where they explain a scientific workflow or its continued availability.

Three distinctions organize the synthesis. First, we separate the origin of data from the locations of computation and action. An AI model trained on ISS images on Earth is different from a model selecting an observation aboard a spacecraft. Second, we distinguish scientific decisions from operational decisions. Choosing an observation that discriminates hypotheses is different from maintaining a thermal set point, although both may be needed for the same experiment. Third, we separate action completion from experimental validity. A pump may execute its command while the delivered volume remains unknown; a spectrometer may return data from an unintended target. Validity depends on whether the evidence answers the intended question under known conditions.

\begin{table}[t]
\centering\small
\caption{A functional taxonomy for interpreting space AI4S claims, organized by complementary capabilities and their evidence requirements.}
\label{tab:taxonomy}
\begin{tabularx}{\linewidth}{P{0.19\linewidth}YY}
\toprule
Function & Scientific contribution & Evidence needed to establish the claim \\
\midrule
Data interpretation & Estimate observables, predict responses, identify patterns or candidate mechanisms & Data provenance, appropriate held-out evaluation, uncertainty and scientific interpretation \\
Scientific decision & Choose a target, setting, intervention or measurement & Explicit objective; acquired evidence changes the choice; meaningful baseline \\
Physical execution & Obtain the selected observation or apply the intervention & Demonstrated action in a specified environment; known human contributions \\
Validity and recovery & Verify the result, diagnose interruption, preserve or restore useful evidence & Checks on sample and measurement state; recovery outcomes beyond cleared alarms \\
\bottomrule
\end{tabularx}
\end{table}

These distinctions also clarify terminology. We use \emph{automation} for execution of defined procedures, which can include feedback, branching, and designed recovery. We use \emph{scientific autonomy} for delegated decisions about acquiring or interpreting evidence within specified objectives and constraints. We use \emph{embodied autonomy} for the capacity to estimate relevant physical state and act on it through sensing, movement, manipulation, or reconfiguration, with outcome verification. Embodiment is functional: a microfluidic device can have a useful restricted action space, while a mobile manipulator can access a broader range of interfaces. The question is which physical capabilities are available, which decisions are delegated, and how their consequences are checked.

\subsection{Central argument and contribution}
Our central argument is that \textbf{space robots can expand the attainable scope of autonomous science when scientifically necessary interventions exceed validated instrument capabilities and timely human support is unavailable}. Many scientific loops can be closed through valves, optical settings, or instrument pointing. Others require access to a new measurement geometry, sample transfer, interface exchange, or preservation after an interruption. The relevant comparison is whether an added capability improves valid scientific return under mission constraints relative to dedicated mechanisms, redundancy, and human-supported operation.

Three questions make the proposition testable: what scientific question is blocked, which observation or physical action is missing, and why cannot the existing apparatus or available support supply it? Scientific decision authority, physical action repertoire, and dependence on external support are separate dimensions. A broader repertoire may enable a new class of experiment without increasing decision authority; a better policy may improve an existing instrument without adding mechanics.

The contribution is an account of this boundary across orbital laboratories and planetary fieldwork. Existing space-laboratory perspectives motivate AI-supported deep-space biology, while terrestrial reviews analyze experimental optimization, laboratory integration, and robotics \citep{bg_sanders2023,grd_tom2024,grd_canty2025,grd_moran2026}. Table~\ref{tab:positioning} locates our synthesis relative to those contributions. We jointly examine scientific function, sample and measurement history, executable actions, external support, and the scientific outcome of intervention. This connects demonstrated capabilities to a choice among competing designs. A worked orbital design, a planetary extension, and matched comparisons in Section~\ref{sec:agenda} make the resulting propositions operational. The search, selection, and source-access limitations of this structured critical narrative review are documented in Appendix~\ref{sec:methods}.

\begin{table}[tbp]
\centering\small
\caption{Position relative to closely related perspectives and reviews, comparing their established emphases with the focus of this synthesis. Record-level source-access limits are documented in the evidence supplement.}
\label{tab:positioning}
\begin{tabularx}{\linewidth}{P{0.23\linewidth}YY}
\toprule
Related work & Established emphasis & Focus of this synthesis \\
\midrule
Deep-space biological laboratories \citep{bg_sanders2023} & AI-supported biological research and self-driving laboratories in deep space & Compare orbital biology with physical sciences and planetary fieldwork using explicit deployment and support boundaries. \\
Terrestrial self-driving laboratories \citep{grd_tom2024,grd_canty2025} & Experimental feedback, integrated hardware and software, scientific acceleration and access & Examine how limited intervention opportunities and experiment history constrain transfer to space. \\
Robotics in self-driving laboratories \citep{grd_moran2026} & Robot functions, instrument interfaces, generality, maintenance and reconfiguration & Connect a missing action to a scientific acceptance condition and compare it with fixed mechanisms, redundancy and human support. \\
\bottomrule
\end{tabularx}
\end{table}

\section{Why space changes the value and requirements of AI4S}
\label{sec:environment}
\subsection{Space as an experimental condition}
Gravity participates in the processes being studied. Reducing buoyancy-driven convection and sedimentation changes how transport and structural evolution can be observed, while biological systems respond to altered mechanical loading and to the combined history of flight exposure. The aim is often to separate mechanisms that are difficult to distinguish under terrestrial conditions, or to test how a mechanism changes when a dominant environmental influence is altered \citep{bg_nasem2023}. A useful AI model must therefore represent the scientific distinction between environments and its consequences for interpreting observations.

The reviewed experiments make this concrete. Materials measurements connect melt transport and growth to models of solidification; foam studies investigate evolution when drainage is suppressed; boiling experiments test the relation between flow morphology and heat transfer \citep{orb_ghritli2021,orb_pasquet2023,orb_mudawar2023ht}. In biology, tissue chips and organism platforms allow controlled measurements that complement whole-organism studies, but each model retains its own limits of interpretation \citep{bio_muscle2023,bio_kidney2024,bio_wormspace2024}. AI can extract difficult observables, connect multiple measurement modalities, and identify candidate explanations. Its scientific value is greatest when those outputs distinguish mechanisms or guide discriminating measurements, rather than merely increasing data-processing throughput.

Planetary in situ work adds a different source of uniqueness: the location and context of the sample are part of the evidence. An outcrop's geometry, neighboring materials, and accessibility constrain its interpretation and how it can be investigated. The act of reaching, touching, drilling, or transferring material changes the measurement opportunity. Target-selection systems and sample-acquisition mechanisms consequently participate in the scientific method itself \citep{pln_francis2017,pln_lawson2025,pln_tagsam2018}. This makes physical autonomy relevant even when the scientific inference ultimately remains with researchers on Earth.

\subsection{Operational conditions across space settings}
An experiment's platform, enclosure, and exposure history determine its operating conditions. An ISS payload can operate inside a pressurized, thermally managed facility with crew access and ground communication. An external instrument faces a different environment. Lunar and martian surfaces have partial gravity rather than orbital microgravity; small-body sampling poses another contact-mechanics problem. Deep-space radiation exposure differs from that in low Earth orbit. Experiments also experience launch, storage, transit, activation, and sometimes return, so the relevant environment is a history rather than a single operating point \citep{bg_nasem2023,bio_ecamsat2020,bg_biosentinel_status}.

The value of local decisions depends on the relation between the time available for a scientifically important decision and the time needed to obtain external assistance. A slowly evolving orbital experiment may remain productively ground supervised, while a transient event or a post-traverse observation opportunity can benefit from local decisions. Interplanetary light time, communications scheduling, and constrained data return make that benefit more pronounced for some tasks \citep{pln_francis2017,rob_braun2026}. Autonomy should be allocated around these task timescales and the support opportunities available in each setting.

\begin{table}[t]
\centering\small
\caption{Setting-specific constraints and proposed design responses for validation. Conditions refer to the experiment's actual enclosure and exposure history.}
\label{tab:environment}
\begin{tabularx}{\linewidth}{P{0.22\linewidth}YY}
\toprule
Setting or constraint & Consequence for scientific work & Design implication to test \\
\midrule
Pressurized orbital laboratory & Microgravity alters transport and loading within a controlled cabin or chamber & Validate mixing, sampling and measurement models; retain crew and ground roles \\
Exposed orbital payload & Vacuum and external exposure differ from cabin conditions; shielding and enclosures modify what reaches the sample & Verify containment, thermal control and calibration under the actual exposure \\
Lunar or martian surface & Partial gravity, destination-specific surface conditions and contact mechanics affect access and sampling & Validate instrument placement and acquisition; retain geological context \\
Small-body sampling & Weak gravity and contact disturbance couple acquisition to material retention & Confirm sample presence, containment and preservation after contact \\
Assistance timing & LEO contact gaps and interplanetary delays impose different limits relative to the decision window & Delegate expiring opportunities locally; retain consultation, pause and escalation options \\
Resources and history: all settings & Finite consumables, restricted access and prior exposure limit which comparisons remain feasible & Track sample history and resources; verify validity and provide feasible exchange or preservation actions \\
\bottomrule
\end{tabularx}
\end{table}

\subsection{The value of learning is constrained by the value of an executable experiment}
Terrestrial AI4S often gains efficiency by choosing trials that reduce uncertainty or improve a target property. In space, the same principle must include the probability that a proposed trial can be executed and interpreted with the resources available. A promising culture condition is of little value if the necessary reagent cannot be delivered or if an unobserved interruption prevents assigning the result to that condition. A high-priority geological target is of limited value if the instrument cannot establish contact or if sampling destroys the context needed for interpretation. Scientific design therefore depends on the available physical capabilities and the observability of their outcomes.

Three forms of value follow from the literature. \emph{Analytical value} comes from extracting more reliable quantities or hypotheses from scarce measurements, as in spaceflight omics and imaging. \emph{Decision value} comes from adapting observations while the opportunity remains available, as in onboard targeting and adaptive acquisition. \emph{Continuity value} comes from maintaining access to valid experiments over time, including diagnosis, preservation, and reconfiguration. The first two already have substantial, though differently distributed, evidence across the reviewed domains. The third often relies on humans, specialized mechanisms, or exceptional recovery efforts \citep{bio_muscle2023,bio_mobp2025,rob_curiosity2022,rob_biolabmaintenance2022}.

These forms of value require complementary validation. Analytical evaluation assesses the quantities inferred from observations; policy evaluation assesses whether acquired observations distinguish scientific explanations; recovery evaluation assesses whether the scientific comparison remains valid after an interruption. The following sections use these distinctions to examine what current systems accomplish and where additional embodied capabilities could make a scientific difference.

\section{AI4S in orbital laboratories}
\label{sec:orbital}
\subsection{Biology: linking inference to the experimental lifecycle}
\label{sec:biology}
Space biology exposes a central discontinuity in current AI4S: the ability to extract increasingly sophisticated hypotheses from flight data has advanced alongside, but often separately from, the ability to acquire those data autonomously. Its literature therefore needs two complementary classifications. One concerns the scientific inference performed by AI, including molecular association, candidate-mechanism identification, and image-based phenotyping. The other concerns the physical experiment: preparing and reviving organisms, maintaining culture conditions, delivering perturbations, acquiring measurements, and preserving samples. Treating either classification alone as the state of space AI4S obscures the missing links between them.

\subsubsection{Molecular inference and reusable data}
GeneLab and the NASA Open Science Data Repository (OSDR) have made spaceflight and analogue data available for reuse beyond the original experiments, while community RNA-seq processing pipelines improve comparability and provenance \citep{bio_genelab2019,bio_osdr2025,bio_rnaseq2021}. They enable researchers to relate sparse experiments across tissues, species, missions, and measurement modalities. These dimensions also introduce heterogeneity: launch conditions, storage, tissue collection, sex, strain, and processing history can affect the signal that a model attributes to spaceflight.

Several complementary computational approaches illustrate what has already been achieved. Knowledge-network embedding links transcriptomic changes in spaceflown mice to broader biological and disease-associated knowledge \citep{bio_network2021}. Graph-based drug-repurposing analysis extends such associations toward candidate interventions \citep{bio_drug2021}. The resulting rankings can organize follow-up work, but a plausible network link is not evidence that a proposed drug prevents a spaceflight phenotype. Symbolic regression provides another route: Li and colleagues use multi-omics and physiological measurements to construct interpretable models of mouse muscle responses, including calcium reuptake \citep{bio_li2023}. Here, the benefit is an inspectable relationship between candidate molecular features and a phenotype.

Cross-study prediction adds a second layer of complexity. Metadata-based feature engineering has been used to predict differential-expression labels \citep{bio_metadata2024}. Harmonized mouse-liver datasets have supported classifiers and comparisons of gene-importance rankings \citep{bio_harmonize2024}. The latter study explicitly discusses limited samples, restricted cohort diversity, and differences in important genes across models with similar predictive performance. These observations matter for experimental autonomy: a high discrimination score alone does not specify which perturbation will resolve a biological uncertainty. Transfer learning can help exploit related terrestrial data; its proposed use in space medicine requires validation in the target population and environment \citep{bio_transfer2023}.

CRISP connects transcriptomics to phenotypic measurements using an ensemble of causal-inference methods \citep{bio_crisp2025}. Its candidate genes are explicitly putatively causal; the paper does not experimentally establish their causal effects. A subsequent retinal application uses molecular features to predict oxidative-damage and apoptosis-associated phenotypes in spaceflown mice \citep{bio_retina2026}.

GLARE provides a more detailed example of how a learning method changes biological interpretation \citep{bio_glare2025}. Sparse autoencoders, including a version pretrained on single-cell data and fine-tuned on spaceflight transcriptomes, supply representations for ensemble clustering. Comparisons with alternative representations and clustering methods assess structure preservation and downstream clustering, rather than treatment efficacy. Applied to the CARA Arabidopsis root dataset, the pipeline recovered the original cell-wall-related findings and identified a hypoxia-associated expression cluster not emphasized in the original analysis. This generates a specific physiological hypothesis for follow-up. Classification of randomly held-out, restructured gene-expression records is not validation on independent missions or plants, and the biological interpretation still requires experimental discrimination. Across these studies, the next step is to connect a candidate mechanism to a feasible perturbation, an appropriate control and an interpretable measurement.

\subsubsection{Phenotyping and local measurement}
Imaging provides a more immediate connection between biological state and possible experimental action. SpaceAnimal supplies annotated imagery of worms, flies, and zebrafish from the China Space Station, supporting computer-vision development and evaluation \citep{bio_spaceanimal2025}. Tracking methods combine appearance and motion information to address occlusion in these constrained observation environments \citep{bio_tracking2025}. A 2026 preprint applies learned pose and behavior analysis to archival ISS rodent video \citep{bio_mousepose2026}. Its discussion of lens soiling, optical aberration, ambiguous views, and human labeling variability is particularly relevant: a model can improve extraction from available images while the measurement process itself continues to deteriorate.

The cited studies principally establish data resources and ground-based analysis. Their prospective control value is to trigger a change in imaging frequency, a diagnostic view or sample preservation when the relevant state becomes observable. Responses to a contaminated view must respect the enclosure: active inspection, sensor redundancy and human assistance may be more practicable than cleaning an animal habitat or sterile chamber.

Measurement hardware is also becoming more compact. A smart-camera platform supports terrestrial \textit{C. elegans} phenotyping and is proposed for spaceflight use \citep{bio_cip2023}; microfluidics-integrated NemaFlex hardware was specifically redesigned around animal loading, size selection, leakage prevention, bubbles, and crew operations before flight \citep{bio_nemaflex2022}. Likewise, nanopore sequencing aboard the ISS demonstrated that sequence data could be acquired in orbit, with simulated real-time analysis and laptop-based assembly on Earth establishing analytical feasibility \citep{bio_sequencing2017}. This is an enabling measurement capability, not evidence of an autonomous sample-to-answer workflow that prepares material, reasons over sequences, and chooses the next biological experiment without human involvement.

\subsubsection{Dedicated culture and assay automation}
O/OREOS used dedicated microfluidic and optical systems to investigate bacterial survival and growth after exposure in orbit \citep{bio_ooreos2011}. EcAMSat compared antibiotic responses in bacterial strains using programmed culture and optical assays \citep{bio_ecamsat2020}. Its report also identifies pre-experiment storage stress as a factor that could obscure small differential effects. These missions demonstrate that meaningful experiments can be performed with tightly scoped automation. They also show why sample history belongs inside the scientific state estimate rather than being treated as incidental logistics.

BioSentinel extended the biological CubeSat approach toward deep-space radiation research \citep{bio_biosentinel2023,bio_biofluidic2023}. Its mission report and NASA operational account distinguish successful payload and radiation-monitoring operation from the absence of yeast growth in the biological experiments, with extended storage before launch identified as a likely contributor \citep{bg_lingam2025,bg_biosentinel_status}. If viability loss is irreversible, sustaining the intended growth assay requires viable backup material or an alternative preparation pathway. Viability testing, redundant samples, preservation methods, and scientific replanning are therefore relevant design priorities. LabSat's combined plant and fungal payload similarly illustrates both the capability of miniaturized dedicated systems and the importance of communication and prelaunch preservation constraints \citep{bio_labsat2025}.

Larger orbital platforms support more elaborate biology. Automated embryonic stem-cell culture on TZ-1 and the WormSpace microfluidic system on the China Space Station demonstrate sustained culture and imaging under predefined experimental procedures \citep{bio_tz2018,bio_wormspace2024}. WormSpace's month-long operation provides evidence for extended dedicated automation; it does not by itself establish learning-driven experimental choice. Human tissue platforms broaden the scientific questions further. Kidney-chip perfusion hardware has flown on two ISS missions, using preloaded media and fixative cassettes to reduce the complexity of crew operations \citep{bio_kidney2024}. Engineered heart tissues support automated functional monitoring followed by molecular analysis \citep{bio_heart2024}. MicroAge develops specialized human muscle culture and stimulation hardware for orbital investigation \citep{bio_microage2026}. These studies establish that carefully engineered interfaces and bounded protocols can already accomplish tasks that would otherwise require extensive laboratory handling.

\subsubsection{What off-nominal experiments reveal}
Two flight studies make the distinction between protocol execution and scientific validity particularly concrete. In a human muscle-chip experiment, automated perfusion and imaging were accompanied by uncertain flow readings, possible bubble or salt effects in the flow sensor, and temperature spikes associated with imaging \citep{bio_muscle2023}. The team ended the planned 14-day experiment after 10 days, initiated preservation from the ground, and relied on crew cold stowage. Usable molecular material was recovered. This was a successful human-supported decision to preserve scientific value under changing conditions. An autonomous successor would need to discriminate sensor error from actual loss of perfusion, estimate the biological consequences of waiting, and choose between continued observation, an approved intervention, and early preservation.

The Modular Open Biological Platform (MOBP) combines programmable cultivation, sampling and preservation with designed power-loss responses \citep{bio_mobp2025}. Its ISS test demonstrated microbial conversion of a plastic-derived substrate, but executed only part of the transfer schedule. The report links prolonged power loss to interruptions and malfunctions involving the NanoRacks BlackBox integration platform; temperatures also exceeded expectations, plausibly because of its cooling fans. These are upstream support problems as well as payload-level uncertainties. Although the program could resume or skip interrupted operations, missing logs and uncertain delivered volumes prevented product-yield calculation. The priority is therefore reliable power, thermal control, containment and state recording, followed by diagnosis of which interface needs intervention. A robot would help only if that interface were accessible and the required action, tools and spares were available.

These cases suggest three priorities for orbital biological AI4S. First, inference should integrate sample lineage and physiological state with machine telemetry. Second, experimental policies should include diagnostic and preservation actions, not only interventions intended to improve a target outcome. Third, hardware should expose verifiable interfaces for the actions that matter: confirmation of transferred volume, cassette identity, containment, imaging quality, and preservation status. A robotic subsystem may provide cassette exchange, inspection, or access to shared instruments, while pumps, valves, and disposable cartridges remain the appropriate means of routine sterile fluid handling. The relevant design unit is the complete scientific workflow, with the division of labor selected from its actual physical requirements.

\subsection{Materials and crystal growth}
\label{sec:orbital-materials}

Microgravity materials research provides a useful distinction between AI that makes an experiment more informative and AI that decides or executes the next experiment. Reduced buoyancy and sedimentation can help separate transport and growth mechanisms, while dedicated furnaces and imaging systems define the accessible conditions. Residual fluid motion and experimental disturbance remain relevant to measurement. The development of in-situ X-ray studies, reviewed by Reinhart et al., also shows why sounding-rocket experiments belong in the methodological lineage but must be distinguished from sustained orbital operation \cite{orb_reinhart2023}.

Two complementary AI approaches have emerged. The first estimates physical parameters from flight measurements. Ghritli et al. combined numerical transport analysis with Bayesian optimization to infer a concentration-dependent diffusion coefficient for GaSb in InSb melt using ISS crystal-growth results \cite{orb_ghritli2021}. Optimization searches a numerical inverse problem against existing observations, producing a transport description relevant to both orbital and terrestrial growth. The accessible abstract supports this parameter-inference result; it does not provide a comparative performance estimate for an adaptive flight experiment.

The second approach extracts structures and trajectories from image sequences. The MASER~13 experiment directly observed equiaxed Al--Cu solidification \cite{orb_murphy2016}; subsequent work by Wu et al. combined learning-assisted radiograph analysis with post-flight tomography and electron backscatter diffraction to examine dendrite growth directions \cite{orb_wu2026}. This is a multiscale interpretation pipeline. A related preprint by Viardin et al. used synthetic phase-field images to train detection of equiaxed dendrites and applied the method to TEXUS~55 observations of a transparent alloy \cite{orb_viardin2022}. Such approaches can recover more information from limited flight opportunities, although weak early features, crowded structures and instrumental artifacts remain relevant to measurement validity. Both examples are sounding-rocket studies.

The comparison identifies a substantive step beyond generic image recognition. A scientific quantity such as growth direction requires linking projected images to material structure and experimental geometry. A prospective controller must then distinguish changes in growth from changes in visibility, overlap or calibration before selecting a new temperature profile. This makes the observation model and the control interface separate validation targets.

Orbital thermophysical experiments provide an important control case. Electromagnetic levitation has supported Si--Ge melt measurements, while ISS electrostatic levitation has enabled measurements of refractory oxides \cite{orb_luo2020,orb_ishikawa2022}. These facilities already embody substantial physical control without thereby constituting AI4S. Ishikawa et al.'s 2022 account describes position feedback and ground-commanded laser heating; it also reports unsuccessful melting of particular oxides associated with unstable charging. The immediate problem in that example concerns sample state and control. Conversely, extending an experimental campaign to unfamiliar materials may expose handling or reconfiguration requirements outside a furnace's built-in action repertoire. The evidence therefore supports evaluating software adaptation, instrument redesign and robotic access separately.

\subsection{Multiphase flow and combustion: prediction, observability and experimental validity}
\label{sec:orbital-fluids}

Flow boiling illustrates a research sequence from flight measurements to physical models and learning-based prediction. ISS experiments characterized heat transfer and critical heat flux in a heated rectangular channel, providing observations against which correlations and mechanistic accounts could be assessed \cite{orb_mudawar2023ht,orb_mudawar2023chf}. Mudawar and colleagues trained separate fully connected feed-forward networks for heat transfer and critical heat flux using consolidated ISS and terrestrial FC-72 measurements \cite{orb_mudawar2024ann}. On randomly held-out test subsets, the reported mean absolute relative errors were 7.99\% for heat transfer and 12.05\% for critical heat flux; the networks outperformed the assessed correlations in the reported database subsets. This is a concrete improvement in predicting measured thermal behavior. A subsequent network addressed pressure drop in the ISS dataset \cite{orb_mudawar2025pressure}. These analyses reuse a flight experimental program rather than constitute additional AI-controlled deployments.

The prediction study randomly partitions database observations into 70\% training, 15\% testing and 15\% validation \cite{orb_mudawar2024ann}. The two error estimates concern held-out observations within the sampled apparatus and conditions, rather than independent flight experiments. Holding out entire runs, geometries or fluids would test transfer beyond the sampled apparatus and conditions. Kinjo et al.'s study explicitly addresses transfer from horizontal-flow data to vertical and microgravity prediction, although its accessible abstract leaves the source platform for all microgravity observations and the deployment location unspecified \cite{orb_kinjo2026}.

An alternative to direct input--output prediction uses learning to expose intermediate physical quantities. Huang et al. analyzed parabolic-flight visualization to connect flow morphology with heat-transfer and critical-heat-flux modeling \cite{orb_huang2024}. Their automated analysis makes the relation between perception and a physical model explicit, while retaining geometric limitations: a two-dimensional observation does not fully reveal three-dimensional wetting, and performance in one channel requires reassessment before transfer.

Experimental dynamics further complicate prospective closed loops. ISS observations of two-phase instabilities motivated additional measurements with altered acquisition settings \cite{orb_mudawar2024instability}. Such work demonstrates the scientific value of revising a measurement protocol, but not AI-directed reconfiguration. Nor should a physical instability automatically be labeled an equipment fault: it may be precisely the phenomenon under investigation. A future decision system must separate a valid unstable state, an inadequate observation, and an apparatus malfunction before using any of them to update a scientific model. This distinction is especially consequential near critical conditions, where optimizing predictive performance alone does not define an acceptable experiment.

Combustion supplies both a longer history and a different observation problem. A 2001 Chinese study combined neural networks and time-series prediction to examine combustion-temperature changes during parabolic-flight microgravity fluctuations \cite{orb_deng2001}. More recently, Gagnon et al. used experimental data to predict flame spread over electrical wires across material, atmospheric and gravity conditions \cite{orb_gagnon2021}. Together, these studies trace neural-network combustion analysis across different experimental settings and scientific observables.

ISS droplet-combustion studies established an instrumented lineage for studying burning and extinction \cite{orb_dietrich2014}. A 2026 preprint applies Segment Anything Model~2, automatic prompting and geometric fitting to extract flame diameters from FLEX-2 images, with manual-reference validation \cite{orb_xu2026}. Earlier observations motivating cool-flame interpretation after visible hot-flame extinction provide an instructive counterpoint \cite{orb_nayagam2012}: a missing luminous contour need not imply the absence of continuing reaction. A prospective adaptive experiment therefore needs an observation policy tied to chemistry as well as reliable image measurements.

\subsection{Soft matter: image analysis within capable dedicated apparatus}
\label{sec:orbital-softmatter}

Foam experiments make the relationship between learning and physical automation unusually concrete. Suppressing drainage permits observations of wet-foam evolution that are difficult to sustain under normal gravity. Pasquet et al. describe ISS measurements supported by a dedicated cell carousel, foam-generation mechanism, cameras and light-scattering instrumentation \cite{orb_pasquet2023}. The apparatus already performs physical operations; AI is introduced into the measurement pipeline. Manual, scripted and neural-network estimates of bubble size are compared, including attention to systematic biases and complementary measurements. This is stronger scientific evidence than reporting segmentation accuracy alone because it asks whether different procedures recover comparable physical observables.

The Bellybutton segmentation method develops this analysis direction using customizable learning from user annotations, including the ISS foam dataset \cite{orb_dillavou2024}. It belongs to the same experimental lineage, not a separate space deployment. Research on hierarchical bubble-size distributions further demonstrates the physical questions supported by these observations \cite{orb_galvani2023}. The scientific contribution combines learning-assisted segmentation with experimental design, independent measurement and mechanistic interpretation.

These studies suggest a prospective route toward adaptive operation: measurement uncertainty could inform whether to continue observing a cell, change an acquisition setting or move to another prepared sample. Existing carousels and dedicated actuators may already cover those actions. A shared manipulator becomes a candidate only for additional exchange or maintenance requirements, with its benefit compared against a dedicated mechanism.

Granular research exposes a further boundary between evidence types. The VIPGran instrument provides configurable, actuated experiments in parabolic flight \cite{orb_aumaitre2018}. Sata et al. subsequently compare machine-learning approaches for predicting dynamical clustering in simulated excited granular gases using related apparatus geometry \cite{orb_sata2025}. The latter is relevant to experimental interpretation, but its validation is numerical. In addition, alternative clustering criteria define different learning targets. Before an algorithm could trigger a new physical experiment, investigators would need to justify which criterion represents the scientific question and validate its estimation from actual observations. Across foams and granular media, the most established contribution is therefore improved measurement and model comparison, while experimental adaptation remains a distinct integration challenge.

\subsection{Cold atoms and the boundary of the robotics argument}
\label{sec:orbital-quantum}

Cold-atom research offers a bounded comparison because experimentally useful controls are often electronic or optical. CAL demonstrated Bose--Einstein condensates in Earth orbit, followed by atom-interferometry experiments aboard the ISS \cite{orb_aveline2020,orb_williams2024}. These are important orbital science results, but they do not by themselves demonstrate artificial intelligence. In particular, ``AI'' in the interferometry literature can abbreviate ``atom interferometer,'' creating a serious retrieval and classification trap. The operational record also matters: the interferometry work describes remote operation and astronaut replacement of a science module. A sophisticated apparatus can sustain experimental sequences while still depending on human intervention for a different class of action.

The relevant AI comparison comes from terrestrial hardware. Wigley et al. demonstrated online learning-based optimization of ultracold-atom experiments, coupling proposed parameter settings to physical trials and measured outcomes \cite{orb_wigley2016}. Anton et al. later benchmarked optimization methods on cold-atom experiments, considering noise and computational overhead alongside optimization behavior \cite{orb_anton2024}. Transfer of these ground-validated loops to CAL would require matching the control interface, objective, computation budget and allowable interventions.

The comparison sharpens the central argument of this review. Scientific decision autonomy can operate through fixed apparatus when the necessary actions are exposed as controllable parameters. Such systems already have physical agency within that action repertoire. The harder question is whether the apparatus can remain scientifically productive when the required action changes category---from tuning an experiment to exchanging, inspecting or repairing its hardware. CAL's documented human intervention identifies this distinction.

Across the physical-science studies examined here, evidence is strongest for learning-assisted analysis of space-generated data, numerical prediction, and well-developed dedicated experimentation. The selected papers do not establish an integrated orbital system that autonomously chooses experiments, handles unfamiliar physical contingencies and validates recovery. The next generation should be evaluated by closing these specific gaps: connect a justified scientific decision to an available action, verify that the resulting observation is valid, and identify when an intervention lies outside the instrument's repertoire. Robotics is most compelling at that last boundary, provided it improves sustained scientific yield over software adaptation, dedicated mechanisms or available human assistance.

\section{AI4S in planetary in situ science}
\label{sec:planetary}

Planetary field science couples interpretation to an unusually constrained measurement process. A promising outcrop may be observable only during a particular traverse; a chemical analysis may consume material; a sample can lose context during collection or transfer. Consequently, the relevant unit of autonomy is a scientific decision together with the observation or intervention that implements it. Autonomous driving is an enabling capability when it makes a scientifically selected location accessible, but distance traversed does not establish improved scientific inference. Conversely, an instrument that adaptively changes a measurement can implement meaningful scientific autonomy without a mobile manipulator or a language model.

The evidence spans several distinct traditions: scientist-configured target selection, adaptive instrument acquisition, model-based opportunistic planning, robotic sample acquisition, and field experiments that redistribute decisions between people and robots. These traditions should be compared by their scientific functions and demonstrated settings. Mission descriptions establish what hardware operated; instrument papers explain measurement mechanisms; autonomy experiments establish which decisions depended on acquired evidence.

This comparison also requires separating three locations: where the data originate, where an algorithm runs, and where the resulting action occurs. Planetary spectra analyzed on Earth are relevant AI4S evidence, but they imply a different response time and authority structure from an onboard controller. The distinctions are retained below even when several publications describe the same mission, so that a new analysis of flight data is not mistaken for a new autonomous deployment.

\subsection{From recognizing a target to choosing a scientific observation}

Early work already combined machine interpretation with physical investigation. Nomad used camera and spectrometer observations with Bayesian classification to identify a meteorite during Antarctic field operations~\cite{pln_nomad2001}. The importance of this example is methodological: a classification depended on obtaining suitable measurements of an actual object. Its scientific action space remained bounded by the instruments, accessible rocks, and classes represented in the investigation.

OASIS made the connection to opportunistic rover science explicit. In tests with FIDO in the JPL Mars Yard, image analysis identified rocks of interest, while a planning component checked whether additional observations could be accommodated. The resulting actions included changing pointing or moving closer for imaging~\cite{pln_oasis2007}. On ground hardware, detection, resource allocation, and acquisition formed a closed sequence. It also identifies an enduring constraint: detecting an interesting object does not make an additional observation feasible within the remaining mission resources.

AEGIS brought target selection into Mars operations, first for Opportunity's Pancam and subsequently for Curiosity's ChemCam~\cite{pln_aegis2012,pln_francis2017}. Rockster segments candidate rocks from images; geometric and visual features are filtered and ranked against scientist-specified preferences before instrument pointing and acquisition. In the reported Curiosity evaluation across 2.5\,km of previously unexplored terrain, preferred-material selection exceeded 93\%, compared with approximately 24\% expected without intelligent targeting, and the selected observations produced geochemical measurements~\cite{pln_francis2017}. This comparison measures delivery of requested material, not discovery accuracy. It establishes the value of using post-drive opportunities before Earth can supply targets. The operational lineage now includes Perseverance's SuperCam: NASA/JPL reports deployment in 2022 and routine geochemical surveying outside the ground-command loop \citep{rev_aegis_status2026}.

These results separate recognition, delivery of requested material and subsequent scientific interpretation. Target profiles make the first autonomous decision tractable, while scientists retain responsibility for whether a familiar lithology or an unexpected observation warrants follow-up. A fair human-targeting comparison must match the imagery and decision time available at selection, rather than use retrospective access to all returned data.

\subsection{Adaptive spectroscopy and the boundary of instrument autonomy}

PIXL on Perseverance adapts acquisition effort using compositional evidence~\cite{pln_lawson2025}. Its onboard rules compare spectral pseudo-intensities with linear classifiers or composition-based thresholds constructed on Earth. A triggered long dwell typically extends a 10-s measurement to 40\,s in total, doubling the expected counting-statistics signal-to-noise ratio. Comparisons of short-dwell spectra with combined short- and long-dwell spectra showed substantive scientific gains: additional counts revised some phosphate major-element estimates and removed spurious trace-element detections, including Zn and Y in a spinel-bearing region. These are demonstrated changes in interpretation, beyond target-ranking success.

The gains were selective. Carbonate measurements in the analyzed scans showed little compositional improvement; many Cu-rule triggers arose from diffraction, and the finite long-dwell budget could be exhausted before later interesting points. PIXL thus demonstrates useful scientific autonomy through a restricted actuator and ground-maintained rules, with performance depending on mineral context and resource allocation. Its 2024 preprint and 2025 journal article describe the same deployment.

The distinction between interpretation and instrument control is particularly important for molecular analysis. MOMA combines laser-desorption and gas-chromatographic mass spectrometry within the planned Rosalind Franklin investigation~\cite{pln_moma2017}. Da Poian and colleagues developed filtering and matching methods using data from a flight-like engineering test unit, together with interfaces through which experts inspect candidate interpretations~\cite{pln_moma2022}. This laboratory evidence concerns the analysis and expert-review stages of the workflow.

NASA describes the intended initial use as processing returned MOMA data on Earth to assist scientists who decide subsequent operations~\cite{pln_moma2024}. Moving from this role to onboard acquisition control would require separate validation of the proposed actions and their scientific consequences.

DraMS extends this instrument lineage toward Titan. Its development description links molecular analysis to DrACO drilling and sample transfer, with a carousel presenting material to complementary analytical modes~\cite{pln_drams2021}. NASA research also considers ML methods for MOMA, DraMS, and future ocean-world mass spectrometers~\cite{pln_scienceautonomy2023}. Integration of these hardware and computational directions remains prospective.

For this family of instruments, a useful future controller would need to distinguish an uninformative spectrum from an unexpected composition, a preparation artifact, or a poorly configured acquisition. The appropriate response could be an additional analytical mode, a reference measurement, or referral to the science team. Classification confidence alone is insufficient to choose among these responses. Scarce sample material makes this distinction consequential: another acquisition may be inexpensive, whereas another preparation step may irreversibly alter the evidence.

\subsection{Sampling and preparation determine what can be inferred}

Sample handling is part of the scientific measurement chain even when its control contains no AI. SAM demonstrates a compact, purpose-built arrangement of sample processing and chemical and isotopic analysis~\cite{pln_sam2012}. The planned ExoMars processing chain similarly connects drilling, crushing, dosing, and presentation to multiple instruments~\cite{pln_exomars2025}. These systems are important comparators to a general robotic laboratory. Their restricted interfaces can make a demanding scientific sequence executable with dedicated mechanisms.

The scientific consequences of preparation depend on the question. Crushing facilitates some compositional measurements while changing spatial relationships. Transfer and storage create a need to preserve identity, source location, and exposure history. These considerations motivate recording the transformation of a specimen rather than treating a successful handover as the end of the task. They also place limits on recovery: repeating a motion cannot restore a lost spatial context or reverse every chemical alteration.

Asteroid sampling illustrates why mechanisms must match the environment. Hayabusa2's sampler was developed for brief contact and collection through a sampler horn~\cite{pln_hayabusa2017}; JAXA subsequently reported its second touchdown sampling operation at Ryugu~\cite{pln_hayabusa2019}. OSIRIS-REx used another specialized architecture, TAGSAM, whose design and qualification explicitly considered the difference between terrestrial tests and asteroid gravity~\cite{pln_tagsam2018}. These are strong examples of scientifically motivated physical capability.

China's Chang'e-5 provides a complementary example of sampling with substantial ground involvement. CNSA's operational account describes drilling, arm-based collection, and sealing, supported by terrestrial reconstruction and simulation of the sampling area~\cite{pln_change2020}. The published control workflow combines environment assessment, planning, image-based refinement, and state-conditioned command execution~\cite{pln_changecontrol2022}. A related study reports deep-reinforcement-learning path planning for precise manipulator control and validation involving the lunar mission~\cite{pln_changeRL2021}. The study supports AI-assisted sampling within a teleoperation workflow; the accessible account leaves the location of training and inference unspecified.

Across these architectures, three interfaces warrant separate assessment: selecting a scientifically justified sample, acquiring material at the intended location, and retaining material suitable for the intended analysis. A platform can succeed at one and fail at another. Autonomous selection becomes scientifically useful only when the execution and verification systems can establish which material was actually acquired. The required verification may include images, container state, instrument response, or a documented chain of custody; the appropriate combination depends on the assay and mission.

\subsection{Planning for information under operational constraints}

Model-based planning connects scientific opportunities to operational constraints. ASPEN represents activities, resources, and operational constraints, while CASPER supports continued modification of a plan as state and goals change~\cite{rob_aspen2000,rob_casper2000}. Such methods explain how a scientific opportunity can be admitted into an executable schedule. Scientific evaluation must additionally assess opportunity detection, measurement calibration, and whether executed actions achieved their intended purpose.

Remote Agent is a useful boundary example. Its architecture integrated model-based planning, execution, and diagnosis~\cite{rob_remoteagent1998}, and a restricted DS1 flight experiment subsequently demonstrated important capabilities and exposed an execution problem requiring ground attention~\cite{rob_ds12000}. This spacecraft-autonomy experiment informs the execution layer of an autonomous laboratory. EO-1's observation-driven planning offers a historical science-loop comparison~\cite{pln_eo12005}, but Earth observation remains background to the present focus on experiments and planetary fieldwork.

Information-driven field research more directly addresses where a rover should measure next. Thompson and colleagues demonstrated geological survey in which spectral interpretation and learned spatial relationships guided exploration near Amboy Crater~\cite{pln_survey2011}. This extends opportunistic detection toward a model of how measurements relate across a site. However, its information value remains conditional on the representation and assumptions used by the planner.

Consequently, route evaluation should report scientific coverage as well as travel cost. It should identify which locations were unreachable, which observations were rejected, and whether adaptive selection changed the resulting sample distribution. The same requirement applies to multi-platform systems: coordination is a scientific contribution when it demonstrably changes the quality, complementarity, or timeliness of evidence.

\subsection{Astrobiology, field analogues, and human--robot comparisons}

Astrobiology exposes the difference between collecting promising material and establishing a biological interpretation. The Life in the Atacama program combined geological and biological instruments with robotic surveys~\cite{pln_zoeoverview2007}, while its science-autonomy experiments investigated adaptive measurement and data return~\cite{pln_zoe2007}. A later study of subsurface microbial habitats used robotic sample collection together with laboratory analysis~\cite{pln_microbes2019}. The reports describe complementary stages of robotic survey, sample collection and laboratory interpretation.

A particularly relevant comparison is the 2025 Zo\"e field study of human-directed, rover-directed and astronaut--rover collaborative exploration~\cite{pln_zoe2025}. The team encoded geological hypotheses; onboard spectral interpretation updated a hypothesis map that informed subsequent exploration. The study reported greater data collection per command cycle under science autonomy and documented changes in geological interpretations as measurements accumulated. This extends evaluation beyond isolated target classification.

Its experimental design also defines the transfer limits. Some analytical inputs came from handheld contact instruments, whose measurements were uploaded to the rover. A shared science team made knowledge transfer between scenarios unavoidable, and some astronaut--rover rendezvous were simulated after operational difficulties. This assisted terrestrial workflow exposes a useful division of labor: onboard analysis and traversal can accelerate coverage while humans provide additional access, measurements and geological context. The relevant next comparison is which of those supporting functions can be automated without degrading interpretation.

The same study reports a direct limit to scientific inference: during the first field season, an incomplete interpretation table omitted a pedogenic origin for repeatedly identified kaolinite, leading to systematically incorrect hydrothermal assessments \citep{pln_zoe2025}. The authors state that another scientist--rover hypothesis-map update cycle would have exposed and corrected the omission. The failure makes hypothesis-representation completeness and revision explicit targets for evaluating scientific autonomy.

Future astrobiology autonomy proposals appropriately emphasize rapid interpretation and prioritization~\cite{pln_astrobiology2025}. Their strongest test will be whether the selected measurements discriminate plausible explanations while preserving enough context for independent assessment. Detection of a pattern associated with biology must remain distinguishable from evidence that excludes relevant abiotic or instrumental alternatives. A negative measurement may constrain a hypothesis; an invalid measurement cannot safely be treated as the same kind of result.

Flight systems have already delegated useful scientific choices within constrained instrument actions. Ground research demonstrates broader adaptive surveys and supports comparisons with human assistance. Sample-return and instrument operations establish the importance of handling, preservation, and verification, but many of these activities remain preplanned or ground supervised. The next research step is to connect these capabilities under explicit scientific and operational authority, while testing whether the integration improves valid evidence rather than merely increasing the volume of autonomous activity.

Table~\ref{tab:space-technical-synthesis} compares methods, scientific results and supporting conditions across eight representative cases. Its first part concerns analysis of acquired flight data; its second concerns implemented experimental or field workflows. Within-study baselines are retained because prediction error, target preference and scientific interpretation measure different achievements. MOBP is explicitly an automation comparator. Together, the cases show both established scientific gains and the specific integration work needed to extend them.

\begin{table}[!htbp]
\centering
\small
\renewcommand{\arraystretch}{1.17}
\setlength{\tabcolsep}{4pt}
\caption{Technical synthesis of representative space-science cases: analysis of acquired flight data. Each row retains its original validation scope and within-study baseline.}
\label{tab:space-technical-synthesis}
\begin{tabularx}{\linewidth}{P{0.15\linewidth}YYY}
\toprule
Case / evidence & Method and scientific task & Demonstrated result & Baseline, support and boundary \\
\midrule
GLARE \citep{bio_glare2025} \newline Ground analysis & Sparse autoencoders, transfer pretraining and ensemble clustering of plant transcriptomes. & Recovered cell-wall-associated patterns and identified a hypoxia-associated cluster for biological follow-up. & Alternative representations and clustering methods; researcher interpretation. Gene-record validation does not establish independent-mission prediction. \\
\addlinespace
ISS crystal growth \citep{orb_ghritli2021} \newline Ground inference & Bayesian optimization of a transport model using existing GaSb/InSb growth observations. & Estimated a concentration-dependent diffusion coefficient. & Experiment--simulation matching; abstract-level access. Comparative predictive benefit and adaptive flight operation are not established by the accessed evidence. \\
\addlinespace
ISS FBCE \citep{orb_mudawar2024ann} \newline Ground prediction & Separate feed-forward networks for heat transfer and critical heat flux. & Randomly held-out test-subset errors of 7.99\% and 12.05\%, respectively; lower than assessed correlations. & Consolidated ISS/Earth FC-72 observations; random 70/15/15 training/test/validation split. Transfer across held-out runs or geometry remains a separate test. \\
\addlinespace
ISS wet foams \citep{orb_pasquet2023,orb_dillavou2024} \newline Post-acquisition analysis & Neural-network segmentation for bubble-size measurements. & Cross-method measurement comparison and bias assessment support physical observables. & Manual/scripted estimates and calibration. Computing location: ? (unreported, 2023); G(i) (inferred ground, 2024 software study). No apparatus feedback established. \\
\bottomrule
\end{tabularx}
\end{table}

\begin{table}[p]
\ContinuedFloat
\centering
\small
\renewcommand{\arraystretch}{1.17}
\setlength{\tabcolsep}{4pt}
\caption[]{Technical synthesis of representative space-science cases (continued): implemented experimental or field workflows. MOBP provides a dedicated-automation comparator alongside AI-supported workflows.}
\begin{tabularx}{\linewidth}{P{0.15\linewidth}YYY}
\toprule
Case / evidence & Method and scientific task & Demonstrated result & Baseline, support and boundary \\
\midrule
MOBP \citep{bio_mobp2025} \newline ISS automation comparator & Programmed fluid transfers and power-loss responses; no AI-guided experiment selection reported. & Microbial substrate conversion; only part of the transfer schedule executed. & No AI comparator. External power/thermal support and missing histories limited quantitative yield interpretation. \\
\addlinespace
AEGIS \citep{pln_francis2017,rev_aegis_status2026} \newline Mars execution & Image segmentation, preference ranking and autonomous instrument targeting. & More frequent selection of requested material than expected without intelligent targeting; successful geochemical acquisitions. & Scientist-defined preferences; blind-targeting expectation. Deployed on Curiosity/ChemCam and Perseverance/SuperCam. Preference agreement is distinct from discovery quality. \\
\addlinespace
PIXL \citep{pln_lawson2025} \newline Mars execution & Ground-built classifiers trigger longer spectral integrations onboard. & Refined composition estimates and rejected some spurious elemental detections. & Short versus combined dwells; benefits vary by mineral and rule. Finite dwell budget; ground rule maintenance. \\
\addlinespace
Zo\"e \citep{pln_zoe2025} \newline Ground field analogue & Spectral interpretation, hypothesis-map updates and adaptive exploration. & Increased data collection per command cycle; evolving geological interpretations. & Shared-team knowledge, handheld inputs and simulated rendezvous limit comparison. Omitted pedogenic kaolinite origin caused first-season hydrothermal misattribution; hypothesis-map completeness requires team revision. \\
\bottomrule
\end{tabularx}
\end{table}

\begin{figure}[p]
\centering
\includegraphics[width=\linewidth]{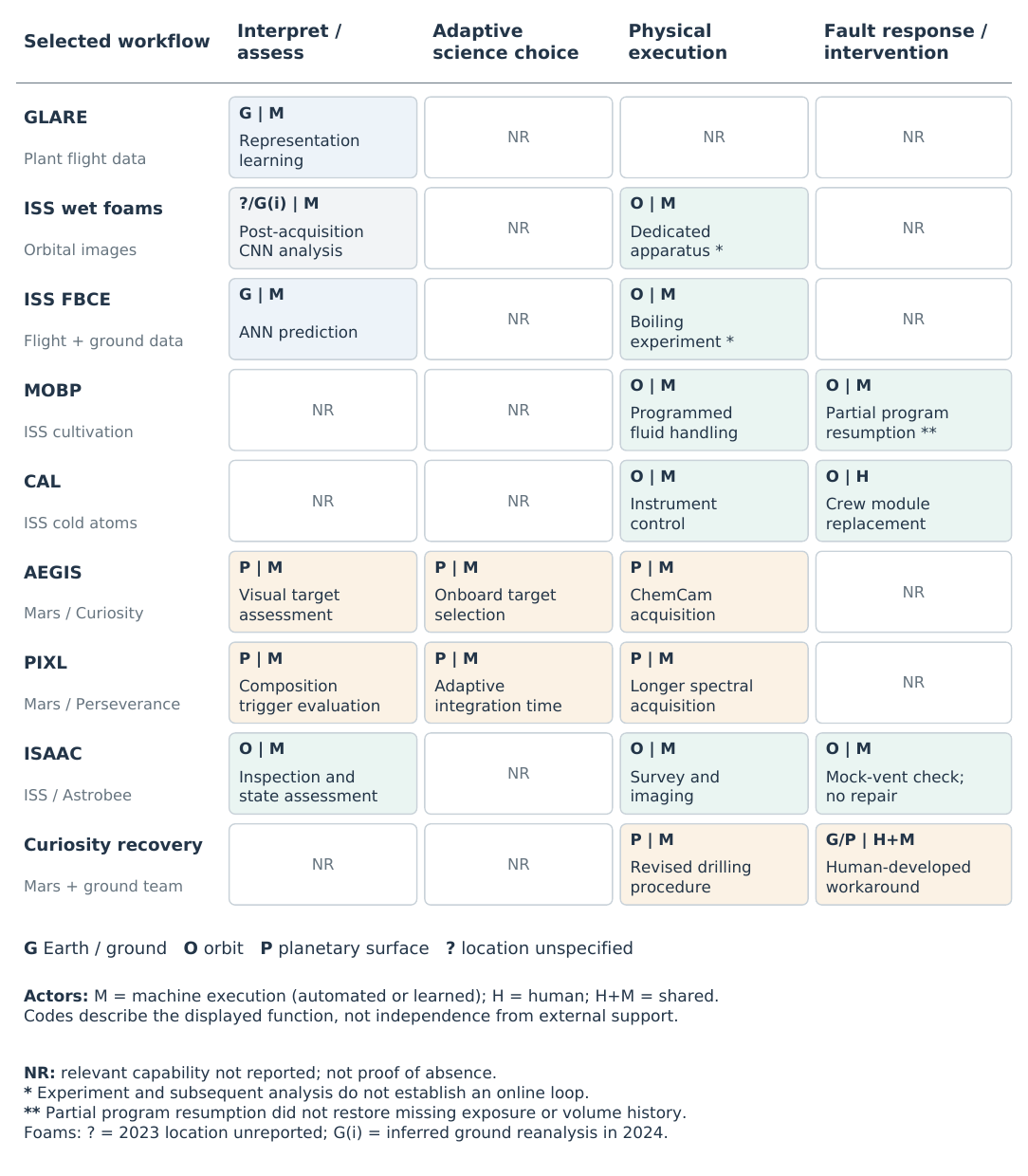}
\caption[Representative evidence by function and demonstrated location]{\textbf{Representative scientific workflows and operational comparators, classified by function and demonstrated location.} Locations and actors are coded independently. M denotes machine execution, including programmed automation; NR denotes a capability unreported in the selected evidence. Analysis location is distinguished from data origin. Interpret/assess denotes machine analysis. Columns show complementary functions across illustrative cases; an integrated loop requires evidence linking those functions. Fault responses distinguish inspection, restart, replacement and restored function. Sources correspond to GLARE \citep{bio_glare2025}; wet foams \citep{orb_pasquet2023,orb_dillavou2024}; FBCE \citep{orb_mudawar2023ht,orb_mudawar2024ann,orb_mudawar2025pressure}; MOBP \citep{bio_mobp2025}; CAL \citep{orb_aveline2020,orb_williams2024}; AEGIS \citep{pln_francis2017,rev_aegis_status2026}; PIXL \citep{pln_lawson2025}; ISAAC \citep{rob_isaac2025}; and Curiosity recovery \citep{rob_curiosity2022}. For foams, ? denotes the unreported 2023 analysis location; G(i) denotes inferred ground reanalysis in the 2024 software study, based on its offline labeled-image workflow. MOBP resumed part of a predefined schedule; missing experimental history remained unresolved. CAL module replacement was performed by crew; fault recovery was not established. ISAAC inspected a mock vent after synthetic telemetry; it did not repair the obstruction. Curiosity's workaround was developed on the ground. AEGIS preferences and PIXL rules were specified on Earth.}
\label{fig:evidence-map}
\end{figure}

Figure~\ref{fig:evidence-map} organizes these cases by demonstrated function and execution location.

\FloatBarrier

\FloatBarrier
\section[Terrestrial AI4S: Progress and Limits of Autonomous Experimentation]{Terrestrial AI4S: Progress and Limits of\\ Autonomous Experimentation}
\label{sec:terrestrial}

\subsection{Parallel routes to experimental autonomy}

Terrestrial laboratories demonstrate three partly independent routes to experimental autonomy: algorithms that choose informative experiments, hardware that executes repeatable operations, and infrastructure that connects instruments, data and people. Reviews already distinguish software and hardware autonomy and identify generalizability, provenance and interfaces as persistent challenges \cite{grd_tom2024,grd_stach2021,grd_canty2026}. The relevant question for space is which combination sustains a particular scientific investigation when physical access and human assistance are restricted.

Robot-scientist research established that explicit hypotheses could drive automated experiments. Early functional-genomics work and Adam connected yeast experiments to machine-generated hypotheses; Adam's conclusions were also confirmed manually \cite{grd_king2004,grd_king2009}. Eve extended this approach towards automated screening and model-guided compound selection, although its economic comparison used existing assay outcomes as a simulated oracle \cite{grd_eve2015}. These studies established scientific decision loops within specified domains.

Experimental design can advance without increasing mechanical complexity. Bayesian reaction optimization evaluates candidate conditions within a defined chemical space \cite{grd_shields2021}. ARES couples nanotube growth and Raman measurements in a fixed apparatus \cite{grd_nikolaev2016}; CAMEO selects diffraction measurements on previously synthesized material libraries, with synthesis outside its demonstrated loop \cite{grd_kusne2020}. Machine learning likewise optimized control parameters in an ultracold-atom experiment \cite{orb_wigley2016}. When the scientific intervention is an instrument setting, adaptive software and reliable instrument control may supply the required autonomy.

The origin of a reported improvement therefore matters. Better acquisition policies can reduce the experiments needed to find a target. Dedicated hardware can reduce turnaround time or variability. A robot can connect previously isolated operations. Longer operating hours can increase throughput without improving sample efficiency. These gains require different baselines and should not be combined into a single acceleration factor. Cost-informed Bayesian optimization also shows how an acquisition policy can account for inventory and purchasing constraints, although its reported evaluation uses existing reaction data rather than a new autonomous hardware campaign \cite{grd_schoepfer2024}. Extending such objectives to scarce flight reagents is a plausible transfer hypothesis.

\begin{figure}[t]
\centering
\includegraphics[width=\linewidth]{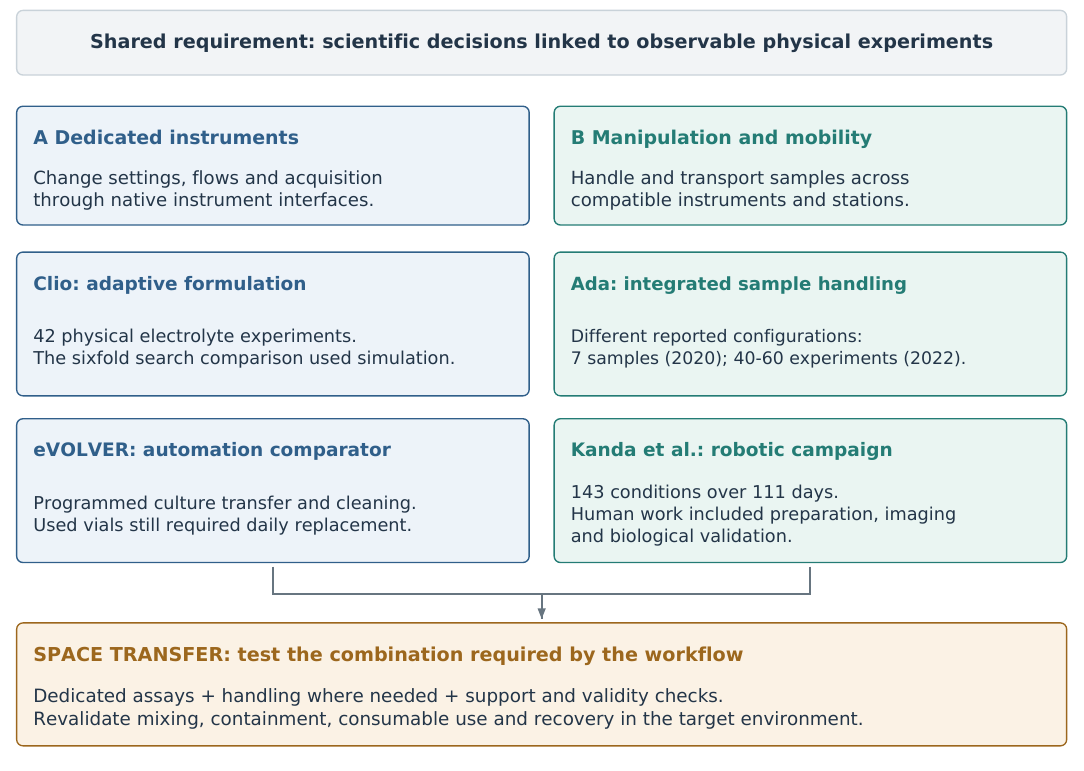}
\caption{Complementary terrestrial routes toward experimental autonomy. Clio combined adaptive formulation with dedicated fluidics; its 42 physical experiments and simulated search-efficiency comparison are distinct results \citep{grd_dave2022}. eVOLVER illustrates dedicated automation with residual vial replacement \citep{grd_wong2018}. Ada reported different replenishment intervals in two configurations and workflows, not a controlled before--after comparison \citep{grd_macleod2020,grd_macleod2022}; Kanda et al. demonstrated a long robotic optimization campaign with specified human preparation, imaging and validation \citep{grd_kanda2022}. The lower panel is an original, proposed space-transfer strategy; the case annotations are not a common acceleration metric.}
\label{fig:parallel_routes}
\end{figure}

Figure~\ref{fig:parallel_routes} summarizes two complementary hardware routes and the support that remained in representative demonstrations.

\subsection{Fixed fluidic systems and hybrid biological laboratories}

Fixed and reconfigurable fluidic platforms are a substantial route to experimental autonomy. Modular flow chemistry couples controllable unit operations to analytical feedback and reaction optimization \cite{grd_bedard2018}. Artificial Chemist applies learning-guided flow synthesis to quantum-dot optical properties \cite{grd_epps2020}. Here, much of the physical complexity is resolved by pumps, valves, reactors and predefined fluid paths. Such systems can change experimental conditions without acquiring the ability to handle arbitrary laboratory objects.

Clio illustrates how this route produces scientific value. Automated mixing and conductivity measurements supplied observations to a Gaussian-process Bayesian optimizer that selected lithium-ion electrolyte formulations. The 42-experiment campaign identified high-conductivity candidates; subsequent pouch-cell tests outside the automated loop supported improved fast-charging performance over a reference electrolyte \cite{grd_dave2022}. Thus optimization of a readily measured property was followed by a separate test of its device-level relevance. The reported sixfold advantage over random search came from surrogate-model simulations, not paired physical campaigns. Rinsing and replicate-reading rules addressed carryover. The transferable components are an informative acquisition policy, a characterized measurement process and separate validation of the target property; their individual contributions should remain distinguishable.

Biological platforms reveal how preventive physical design can substitute for general manipulation. eVOLVER integrates culture sensing, environmental control and fluidic multiplexing. Its automatic transfer between vials and fluid-path sterilization addressed biofilm formation, although used vials still required daily replacement in that experiment \cite{grd_wong2018}. Chi.Bio combines culture measurements with media delivery, illumination, stirring and temperature control \cite{grd_steel2020}. These systems are important automation comparators; ordinary feedback control should not be relabelled as AI solely because it operates without continuous human commands. Their relevance is that scientific feedback can be physically embodied in compact dedicated devices.

Hybrid platforms extend this approach when a measurement cannot be integrated into the vessel. ReacSight connects bioreactor arrays to cytometry through automated sampling and supports reactive biological control. Its stated consumable configuration allowed 192 samples before human intervention \cite{grd_bertaux2022}. BioAutomata links Bayesian optimization to robotic construction and screening of genetic designs; its authors explicitly identify the need for additional quality control when assumptions about reliable DNA assembly cease to hold \cite{grd_hamedirad2019}. Thus an apparently closed design--build--test loop can still depend on unverified intermediate operations.

These platforms are relevant to orbital culture experiments, but terrestrial success does not establish direct hardware portability. For example, Chi.Bio's measurement sequence allows liquid to settle after stirring \cite{grd_steel2020}. A flight implementation would have to establish that sampling, mixing, gas exchange and optical measurements remain valid in its gravity and containment regime. Reviews of autonomous microfluidics also identify blockage, fouling and integration as continuing concerns \cite{grd_damir2026}. A fluidic architecture therefore requires observable fluid paths, replaceable modules and validated cleaning procedures.

\subsection{What manipulators and mobile robots add}

Robotic manipulation becomes valuable when the scientific workflow crosses instruments or requires heterogeneous sample handling. Ada automated solution preparation, film deposition, annealing and characterization. Its initial organic thin-film workflow required consumable replenishment after seven samples. A later combustion-synthesis workflow, with additional handling and characterization capability, reported approximately 40--60 experiments before replenishment \cite{grd_macleod2020,grd_macleod2022}. These are capacities of different configurations, not a controlled estimate of the benefit of added robotics. The later study's multiobjective Bayesian optimization used expected hypervolume improvement to resolve trade-offs between film conductivity and annealing temperature. That scientific decision capability is distinct from replenishment capacity: a larger stock of consumables can prolong a campaign without making its experimental choices more informative.

The mobile robotic chemist of Burger et al. conducted 688 photocatalysis experiments over eight days, using a mobile manipulator to operate multiple stations within a researcher-defined search space. Its sixfold improvement concerned catalyst activity, rather than laboratory productivity \cite{grd_burger2020}. Mobility expanded access to existing instruments; it did not remove the need to define chemical hypotheses, provision materials or establish feasible operations. Similarly, robot-assisted flow synthesis combines AI route planning with physical module reconfiguration, while experts supply implementation details \cite{grd_coley2019}. Fixed fluidics and manipulation are therefore complementary components, rather than mutually exclusive stages of progress.

Other integrations broaden the scientific loop. Synbot connects synthesis planning, hardware execution and analytical feedback, but still requires periodic replenishment and waste removal \cite{grd_ha2023}. Multiproperty molecular discovery joins predictions to synthesis and measurements \cite{grd_koscher2023}. Dai et al. combine mobile transport with complementary analytical instruments and rule-based follow-up experiments; identifying an unexpected product nevertheless required additional human interpretation and characterization \cite{grd_dai2024}. Transporting samples to an instrument is different from establishing what its output means scientifically.

The distinction is especially clear in long biological campaigns. Kanda et al. used a dual-arm platform to optimize cell differentiation over 143 conditions across 111 days. The methods specify human preparation and loading, sampling, image acquisition and downstream biological validation \cite{grd_kanda2022}. The campaign demonstrates productive robotic search, but its calendar duration is not a measure of uninterrupted autonomy. For space transfer, preparation and scheduled operations within the experimental loop must be separated from unplanned recovery and subsequent expert validation: they impose different requirements on crew access, automation and communication.

\subsection{Negative results, invalid evidence and operational failure}

An autonomous laboratory must distinguish at least three outcomes. A valid experiment that fails to produce the desired material is a scientific negative result, provided execution and measurement checks support that interpretation. An unreliable measurement or incorrect interpretation can invalidate a particular scientific claim. A failed transfer, blocked line or unavailable instrument is an operational failure. Treating all three as a low objective value can teach an optimizer the wrong relationship between an intervention and its outcome. Conversely, treating every undesired result as a hardware fault can suppress informative observations. This distinction is an analytical requirement for space transfer.

A-Lab exposes these different boundaries in one workflow. Literature-informed recipes and thermodynamics-guided active learning linked synthesis outcomes to follow-up precursor choices. The corrected publication reports 36 synthesized targets among 57 attempted, using 353 experiments over 17 days; its correction also revises phase-assignment and novelty claims \cite{grd_szymanski2023,grd_szymanski2026}. Manual cleaning, replenishment and hardware-exception handling remained part of operation. Subsequent human regrinding and higher-temperature heating produced two additional targets. The authors place these treatments outside the scope of the active-learning algorithm then used, rather than demonstrating that new robot hardware was required \cite{grd_szymanski2023}. The installed platform already performed heating, grinding and transfer, but this does not establish that intermediate regrinding was integrated into an executable autonomous sequence. Three limitations must therefore be separated: operations admissible to the scientific policy, their integration into a workflow, and the physical capabilities of the equipment. This case supports extending and validating the first two before attributing the bottleneck to missing mechanics. The correction belongs to the same experimental lineage and adds no independent demonstration of reliability.

An expanded action repertoire is useful only when the system can decide whether an action succeeded and whether another action preserves the experiment. Early Chemputer development provides an instructive alternative to general visual reasoning: optical phase-boundary sensing was unsuitable for some coloured or cloudy mixtures, and conductivity sensing supplied a workable solution \cite{grd_steiner2019}. This was engineering redesign, not autonomous self-repair. It nevertheless shows why an extra sensor or a constrained interface can be preferable to a more elaborate general-purpose controller.

Recent robotic work begins to address execution-state feedback directly. LIRA couples localization and visual inspection to predefined correction of selected translational placement errors. Its 97.5\% result on 200 deliberately introduced test cases measures inspection and reasoning accuracy, not physical recovery success. More complex errors can trigger human intervention, and integration with a complete scientific decision loop remains a stated next step \cite{grd_zhou2025}. The LabRobFail preprint separately evaluates failure analysis and downstream recovery in simulation \cite{grd_wang2026labrobfail}. Evidence of useful recovery must additionally show that sample identity, contamination status and subsequent measurements remain acceptable.

Maintenance and reconfiguration are explicit research directions in laboratory robotics. Moran et al. identify serviceable hardware, tooling, labware management and orchestration as research prerequisites for reducing human support between campaigns \cite{grd_moran2026}. The ADePT framework separately distinguishes adaptability, dexterity, perception and task complexity \cite{grd_salazar2026}. These agendas identify capabilities to investigate; selecting a space intervention still requires a documented task bottleneck and evidence that the intervention preserves useful scientific outcomes under the available support policy.

\subsection{Protocols, orchestration and language-model agents}

Reproducible execution depends on more than choosing the next experiment. Chemputer-style programs and subsequent synthesis-database validation make operations explicit and executable \cite{grd_steiner2019,grd_rohrbach2022}. Compact dedicated platforms further show that multiple synthesis procedures can be implemented without a mobile laboratory robot \cite{grd_manzano2022}. Their contribution is a structured connection between chemical procedures and physical resources; a stored protocol by itself is not scientific autonomy.

Cross-platform work with the chemical description language demonstrates both the benefit and the remaining gap. Rauschen et al. transferred programs across four hardware configurations in two laboratories, but differences in materials, drying and crystallization still required investigation, and some operations remained manual \cite{grd_rauschen2024}. Software portability does not ensure that two devices realize the same physical process. Computer-science abstractions provide structures for recording that process \cite{grd_canty2024}; Canty et al. further identify accessible instrument interfaces, experimental metadata, calibration and study-specific quality controls as requirements for trustworthy SDL integration \cite{grd_canty2025}. In a flight implementation, these requirements imply that a replacement cartridge or changed procedure must retain its material identity, configuration and calibration history, so that any change to the scientific comparison can be assessed.

ChemOS organizes planning, instruments, analysis and data exchange through modular interfaces \cite{grd_roch2020}. Coscientist and ChemCrow add language-mediated access to documentation, code and chemical tools \cite{grd_boiko2023,grd_bran2024}. Their contribution is language-mediated translation of scientific intentions into supported tool calls; physical execution and fault response require separate observation and validation.

AutoLabs makes the validation distinction explicit. Its evaluation separates procedure structure and chemical quantities from hardware execution. All five expert-led test procedures were run in instrument simulation, but only the first two were physically executed; human verification remained part of the workflow \cite{grd_panapitiya2026}. For disconnected operation, the executable action set, local model availability, validation rules and fallback authority would require mission-specific testing.

\subsection{Conditions for transfer to space}

The terrestrial routes suggest three distinct transfer decisions. If the required change is an experimental setting, improved decision-making can operate through existing instrument controls. If it is a new sequence of supported operations, protocol and orchestration changes may suffice. If a necessary physical operation remains inaccessible, a dedicated mechanism, replaceable module, shared arm or mobile robot becomes a candidate. Hardware should be added only after identifying which of these limits constrains the scientific task. One candidate flight architecture pairs shared handling for occasional exchange with dedicated devices for repeated measurements.

Transfer also requires revalidating the relationship between observations and scientific claims. A proxy must remain informative under the flight measurement and containment conditions. Intermediate states must be observable enough to distinguish relevant scientific variation from execution errors, or the system must retain unresolved uncertainty. Recovery procedures must preserve sample traceability and identify which comparisons survive an interruption. Support records must separate setup, routine replenishment and calibration, unexpected intervention, and subsequent scientific assessment. Missing intervention reports cannot be treated as evidence of unattended operation. Existing SDL discussions of calibration, metadata and partial success make this reporting problem explicit \cite{grd_canty2025}.

The appropriate comparison is between complete workflows under matched scientific goals, resources and support opportunities. Fixed automation, software-level adaptation, a specified robotic intervention and feasible human assistance are alternatives with different costs. To identify the value of manipulation, the compared systems should have equivalent sensing and decision policies where practicable, or report those differences explicitly. Measurements should distinguish informative experiments, throughput, support burden and validity after interruption, including preserved samples and justified stops. Section~\ref{sec:agenda} develops this comparison into testable designs. The terrestrial lesson is to match scientific choices, executable procedures and verified physical capabilities.

\section{Space robotics as an enabler of scientific autonomy}
\label{sec:robotics}

The case for robotics begins with a missing scientific action. An experiment may need a specimen repositioned, a reference presented to a sensor, an inaccessible condition inspected, or a damaged workflow bypassed. Dedicated mechanisms, shared manipulators, mobile platforms, and human teleoperation offer different ways to supply these functions. Their value depends on whether the action can be verified and whether its benefits exceed additional resource use, disturbance, and failure opportunities. The relevant requirement is therefore a justified physical intervention repertoire.

\subsection{Observing the state that makes an experiment valid}

An experimental controller needs more than the command history. A completed motion does not establish sample presence, a sealed container, or correct contact with a surface. Perseverance's early coring operations illustrate the importance of observing the contents of a sample tube before treating acquisition as confirmed~\cite{rob_perseverance2021}. The broader implication is that sensing should be selected around the scientific failure modes: an image may establish geometry while leaving chemical contamination unresolved, and a motor-current trace may establish resistance while leaving sample identity unresolved.

Astrobee offers a mobile sensing platform within the pressurized ISS~\cite{rob_astrobee2016,rob_astrobee}. Its relevance to scientific facilities lies in access to views and locations that fixed instrumentation may not cover. ISAAC demonstrated onboard-planned surveys with two Astrobees and an inspection response to synthetic anomaly telemetry and a mock obstructed vent~\cite{rob_isaac2025}. The flight demonstration covered observation and state assessment; cargo transfer remained simulated, and no manipulator cleared the mock obstruction.

This distinction suggests a useful progression of evidence: detect an anomaly, localize its physical cause, establish which responses are feasible, execute an authorized response, and verify the resulting experimental state. Success at an earlier step does not establish the later steps. For scientific platforms, the final check must include the sample or measurement conditions, rather than merely disappearance of an alarm.

\subsection{Manipulation, specialized mechanisms, and scientific interfaces}

Existing space facilities already contain substantial physical automation. Biolab combines an automated section with crew-operated functions and ground telescience~\cite{rob_biolab,rob_biolab2013}. The arrangement demonstrates useful sample handling without a general-purpose mobile laboratory robot. Likewise, planetary sample systems connect specialized acquisition mechanisms to restricted containers and analytical interfaces~\cite{pln_sam2012,pln_exomars2025}. These are substantive alternatives against which added manipulation should be evaluated.

The advantage of a shared arm would be the ability to support several instruments or contingencies with the same hardware. Its disadvantage is that instruments, containers, contact conditions, and contamination boundaries must all be compatible with the arm's verified skills. A nominal capability such as grasping is therefore too coarse for an evidence map. Exchanging a capped cartridge, handling an exposed biological sample, and reconnecting a fluidic interface impose different requirements for containment, state estimation, and post-action validation.

Chang'e-5 demonstrates how accurate manipulation can be distributed across planning, visual measurements, flight mechanisms, and ground operations~\cite{pln_changecontrol2022,pln_changeRL2021}. It provides a concrete example of matching automation to individual bottlenecks. Where predictable geometry permits a dedicated carousel or transfer mechanism, that design may provide the needed science capability with a smaller validation burden than general manipulation.

\subsection{Recovery is bounded by hardware and scientific reversibility}

Curiosity's drill-feed failure supplies unusually instructive operational evidence. Ground and flight investigations eventually enabled feed-extended drilling, with the first recovered sample acquisition reported 521 sols after the initial stall~\cite{rob_curiosity2022}. The rover's existing arm and drill capabilities provided an alternative way to perform the scientific operation. The recovery demonstrates physical flexibility through ground-developed operational redesign, with substantial human effort.

InSight illustrates a limit of physical adaptability. The deployment arm supplied friction and direct support that assisted parts of the HP3 mole's penetration attempts, but the probe did not reach the depth required for the intended heat-flow measurement. The analysis identifies a soil--probe interaction outside the assumptions supported by the tested configuration~\cite{rob_insight2022}. Available interventions produced partial progress without supplying sufficient sustained penetration. The case motivates validation of hardware--environment assumptions and records a scientific objective that remained unmet despite additional physical support.

OSIRIS-REx separates acquisition from preservation. Material obstructing the collector's retention flap prompted a stowage operation that NASA describes as requiring continuous ground oversight and input over two days~\cite{rob_stow2020}. The action protected acquired evidence, but its successful completion was a human-supported operational response. Scientific recovery can therefore mean retaining material or terminating a risky sequence, rather than resuming the nominal program.

Orbital maintenance has equally concrete prerequisites. A Biolab maintenance account describes an astronaut replacing centrifuge drive belts with the appropriate equipment and support~\cite{rob_biolabmaintenance2022}. The lesson is that maintenance depends on accessible interfaces, tools, spare parts, and a practicable procedure. A robot cannot replace a part that the mission did not carry, or reach a mechanism behind an enclosure that was not designed for intervention. RELL's use with an external robotic system illustrates instrumented infrastructure diagnosis~\cite{rob_rell}; it remains an indirect contribution to scientific availability, not a scientific experiment.

After any intervention, operational and scientific recovery should be assessed separately. Re-establishing flow may not restore the intended exposure duration; reseating a sample may alter its orientation; transferring a specimen may break comparability with earlier measurements. These proposed evaluation requirements derive from the workflow. The system should preserve a record of the interruption and permit a conclusion that a run is invalid even when the hardware is functioning again.

The orbital biological cases in Section~\ref{sec:biology} identify preservation and termination as meaningful alternatives to continued throughput \citep{bio_muscle2023,bio_mobp2025}. MOBP's upstream power and thermal interruptions also require diagnosis to distinguish facility conditions from instrument, controller, or manipulation faults before selecting an intervention.

\begin{figure}[tbp]
\centering
\includegraphics[width=\linewidth]{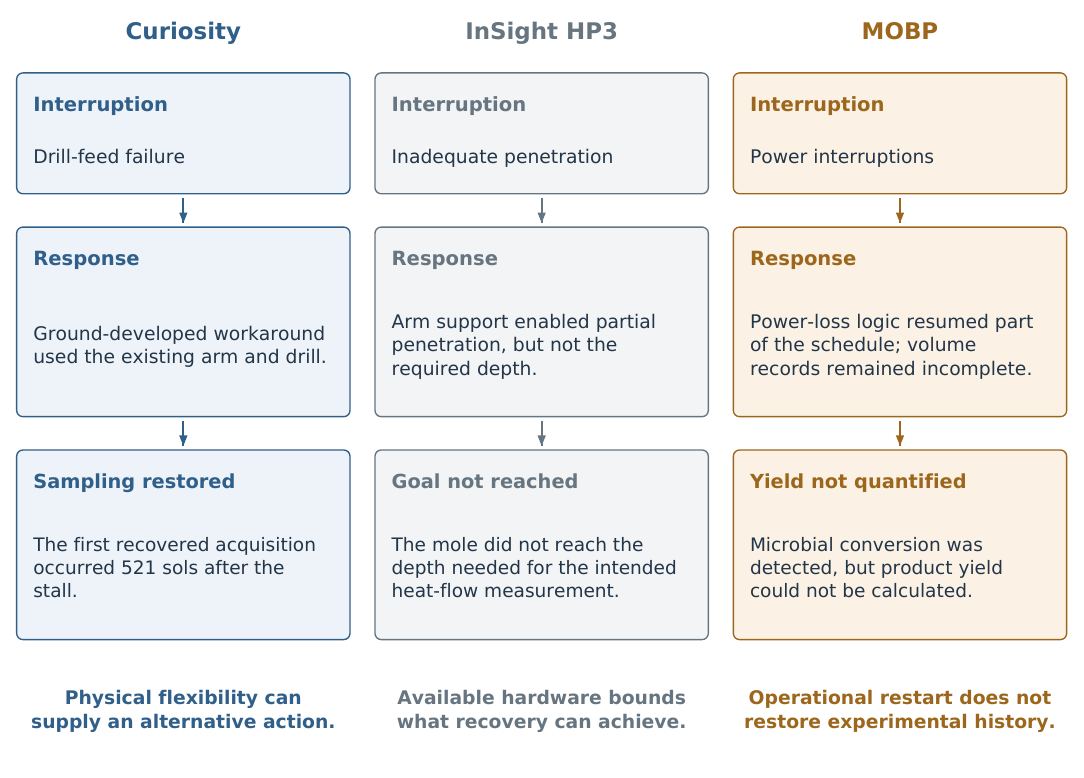}
\caption{Different outcomes of attempted recovery. Curiosity recovered sample acquisition through ground-developed feed-extended drilling \citep{rob_curiosity2022}. Arm assistance helped parts of InSight's HP3 penetration attempts, but the mole did not achieve the required depth \citep{rob_insight2022}. MOBP partially resumed programmed operations after interruptions, but uncertain liquid volumes and incomplete records prevented calculation of product yield \citep{bio_mobp2025}. These cases distinguish physical adaptability, bounded mechanical benefit, and loss of quantitative scientific validity.}
\label{fig:recovery_cases}
\end{figure}

Figure~\ref{fig:recovery_cases} contrasts three outcomes that an autonomy claim must keep separate: restored scientific access, an unmet physical requirement, and resumed operation with incomplete quantitative evidence.

\subsection{Connecting science goals to executable actions}

Model-based autonomy provides tools for linking goals, resource constraints, and execution feedback. Remote Agent, ASPEN, and CASPER demonstrate important elements of this tradition~\cite{rob_remoteagent1998,rob_ds12000,rob_aspen2000,rob_casper2000}. Their relevance is architectural: plans must be revised when the world differs from expectations. However, computational plan repair must not be described as physical fault repair. A new feasible command sequence may still lack the calibration, controls, or provenance needed for a scientifically comparable result.

OASIS and PIXL show two ways to restrict the connection between evidence and action~\cite{pln_oasis2007,pln_lawson2025}. One admits an extra observation after checking resources; the other changes acquisition effort within an instrument. For more complex experiments, an analogous interface would expose authorized actions, their preconditions, the evidence needed to confirm completion, and the conditions under which the system must stop or escalate. A planner's confidence cannot substitute for a missing physical observation of a critical precondition.

The future integration problem is thus not simply whether one controller can command multiple devices. It is whether each scientifically meaningful transition has a reliable physical interpretation. The controller needs to know which transformations are reversible, which consume scarce material, and which change the meaning of later observations. Those properties should constrain experimental planning before execution rather than be reconstructed after a failure.

\subsection{Transfer requires matching physical and operational conditions}

Earth demonstrations are essential but validate a particular environment. EELS demonstrates adaptive task and motion planning for a snake-like platform in terrestrial settings relevant to icy-world access~\cite{rob_eels2024}. The demonstrated capability is terrestrial access; extraterrestrial operation and assay integration require separate validation. Access, sample acquisition, and interpretation remain distinct parts of the scientific chain.

Environmental distinctions also change the viable mechanism. An ISS cabin free-flyer uses an atmosphere and operates among crew and equipment; its design cannot simply be exported to vacuum. Asteroid sampling has contact and material-transport conditions unlike a terrestrial bench~\cite{pln_tagsam2018}. DraMS/DrACO development explicitly treats a cold sample path and a warmer analytical interior~\cite{pln_drams2021}. These examples argue for environment-specific interfaces and tests.

Relevant validation should therefore combine realistic interfaces and materials with the dominant environmental factors. Testing must also represent the available support: a field robot repaired by nearby engineers after each run has not demonstrated months of operation without access. Conversely, a ground-supported flight system can still provide valuable science. The experimental description should report the support policy, its timing and the work supplied by people.

\subsection{Human assistance and simpler designs remain comparators}

ESA's Analog-1 experiment connected an ISS astronaut to an Earth-based rover performing geological sample collection with visual and force feedback and advice from a geological team~\cite{rob_analog2019}. The operator was in orbit, but the robot and rocks were on Earth. This distributed teleoperation experiment supplies a concrete human--robot alternative for sample collection under constrained access.

Field comparisons between human-directed, autonomous, and collaborative exploration further show why the unit of comparison should be the whole investigation~\cite{pln_zoe2025}. The communication delay, amount of expert work, scientific objectives, and accessible action set must be specified. A recent deep-space autonomy review likewise links the need for local decisions to the relation between mission timescales and Earth response times~\cite{rob_braun2026}. That reasoning supports autonomy where waiting is costly, while leaving room for human consultation where the task permits it.

Table~\ref{tab:interventions} makes this comparison at the level of the experimental bottleneck. Additional manipulation is a candidate when the needed intervention lies outside the dedicated apparatus's repertoire; an undesired result or uncertain measurement alone does not establish that requirement.

\begin{table}[tbp]
\centering\small
\setlength{\tabcolsep}{4pt}
\renewcommand{\arraystretch}{1.16}
\caption{Choosing an intervention from the experimental bottleneck. Scientific negatives, invalid evidence, and physical limits require different responses. Cited cases establish the bottlenecks; candidate pathways and acceptance conditions are proposed here for evaluation. Dedicated automation and human assistance remain explicit comparators.}
\label{tab:interventions}
\begin{tabularx}{\linewidth}{P{0.23\linewidth}YYY}
\toprule
Experimental bottleneck & Dedicated or human-assisted alternative & Possible robotic contribution & Scientific acceptance condition \\
\midrule
\textbf{Policy or workflow limit.} A useful synthesis treatment lies outside the implemented policy or workflow \citep{grd_szymanski2023}
& Expand the experiment policy or integrate additional steps using existing equipment.
& Add mechanics only if the revised protocol requires an unavailable physical operation.
& Establish whether policy, workflow or hardware is limiting; retain valid negatives. \\[4pt]
\textbf{Unreliable evidence.} Ambiguous signal or phase assignment \citep{grd_steiner2019,grd_szymanski2026}
& Use an appropriate sensor or independent analysis; audit interpretation.
& Inspect or reposition only where physical geometry causes the ambiguity.
& Establish why the observation or interpretation is now credible. \\[4pt]
\textbf{Routine physical support.} Fouling or depleted consumables \citep{grd_wong2018,grd_macleod2022}
& Programmed cleaning and fluid transfer; human vial replacement or replenishment.
& Candidate module exchange where fixed fluid paths cannot provide the required action.
& Preserve cleanliness and sample lineage; include replenishment and support effort. \\[4pt]
\textbf{Workflow access gap.} Samples must cross instruments \citep{pln_sam2012,grd_dai2024}
& Dedicated carousel or transfer path; human handling and interpretation.
& Transport samples among compatible synthesis and analytical stations.
& Verify identity, transfer completion and analytical comparability. \\[4pt]
\textbf{Execution fault.} Known sample-placement error \citep{grd_zhou2025}
& Candidate fixtures and presence checks; human response to unsupported errors.
& Bounded placement correction followed by reinspection.
& Measure actual correction and subsequent experiment validity, not detection alone. \\[4pt]
\textbf{Physical capability limit.} Failed actuation or unsuitable contact mechanics \citep{rob_curiosity2022,rob_insight2022}
& Built-in alternatives or redundancy where available; ground-developed procedures.
& Exploit existing arm or tool capabilities; missing mechanics can still prevent recovery.
& Demonstrate recovered science; retain failed attempts and the human support burden. \\[4pt]
\textbf{Validity lost across restart.} Exposure or volume history is uncertain \citep{bio_mobp2025,bio_muscle2023}
& Preserve interruption records; humans authorize preservation or restrict analysis.
& Candidate preservation or transfer; manipulation cannot reconstruct missing history.
& Re-establish comparability or exclude affected quantitative claims. \\
\bottomrule
\end{tabularx}
\end{table}

Robotics is valuable when it supplies a needed intervention that cannot be provided adequately by fixed mechanisms, electronic reconfiguration, redundancy, or the available human support. Evaluate each added intervention by its contribution to valid observations, sample preservation, recovery, and support demand, accounting for mass, power, contamination pathways, and verification effort.

\section{Toward sustained, embodied scientific campaigns}
\label{sec:agenda}
\begin{figure}[t]
\centering
\includegraphics[width=\linewidth]{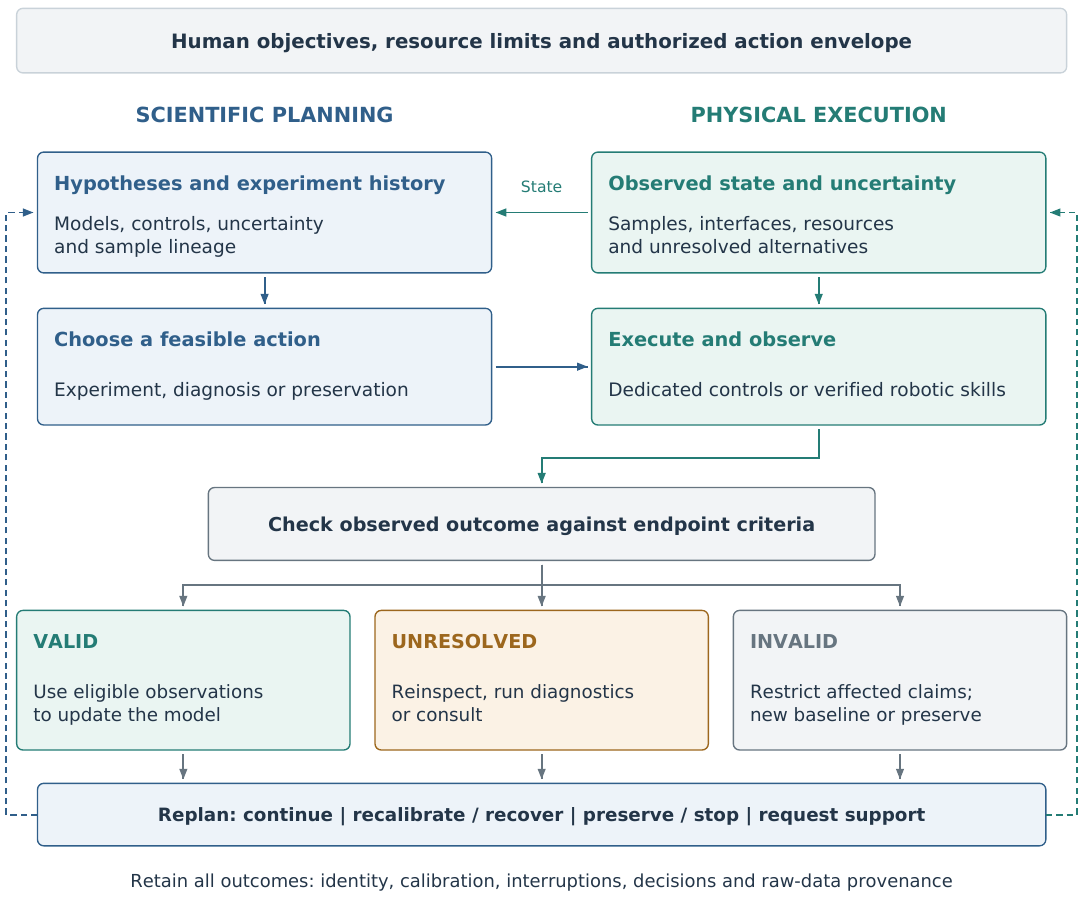}
\caption{Proposed integration of scientific decisions, physical execution, and validity checks. Scientific planning selects an executable action, while physical-state estimation and observation establish the status of the experiment. Valid, unresolved and invalid outcomes lead to distinct responses before replanning. Invalidity is specific to the affected scientific claim; all outcomes and interventions remain in the record. Dashed returns update both accounts of state. Continuation, recovery, preservation, stopping and requests for support are explicit policy choices. Original schematic based on the synthesis in Sections~\ref{sec:biology}--\ref{sec:robotics}.}
\label{fig:architecture}
\end{figure}

The next generation of space AI4S should be evaluated through the scientific work it can sustain and the questions it can newly address. Bounded adaptive measurements establish useful components. Figure~\ref{fig:architecture} and the worked orbital design and planetary extension below specify observations, actions, scientific acceptance conditions, and comparators grounded in reported bottlenecks. These are prospective designs, not flight-validated systems.

\subsection{A coupled scientific and operational state}
A scientific controller needs two connected accounts of the world. One describes hypotheses, models, controls, uncertainties, and the information expected from a measurement. The other describes samples, instruments, consumables, geometry, exposure history, and the health of the measurement process. Neither can be inferred completely from the other. A change in an optical signal may indicate a biological response, a bubble, an illumination change, or a contaminated view. The controller should retain these alternatives until observations distinguish them.

This suggests maintaining an explicit belief over both scientific and operational state. A candidate action may then be an experiment, a diagnostic measurement, a physical intervention, a preservation step, or consultation with Earth. The appropriate choice depends on expected scientific benefit, the likelihood of obtaining a valid observation, resource use, and the consequences of losing an irreplaceable sample. This formulation extends the resource-aware planning seen in OASIS and the constrained acquisition changes seen in PIXL toward broader experimental workflows \citep{pln_oasis2007,pln_lawson2025}. Specialized estimators and planners may be easier to validate and maintain.

The belief state must be tied to available observations. For each scientific endpoint, the design should specify admissible sample histories, calibrated measurements, and known ambiguities. A command acknowledgement establishes a commanded transition, not necessarily delivered volume, specimen identity, or contamination state. If several histories remain compatible with the telemetry and imply different scientific interpretations, the endpoint remains unresolved. Additional sensing may resolve the ambiguity without manipulation; an intervention may instead destroy evidence needed for diagnosis. This ordering makes state verification a design requirement with observable inputs, rather than an assumed validity oracle.

\subsection{Interfaces designed for verification and intervention}
Robot-enabled laboratories require scientific interfaces as much as better manipulators. A useful cartridge interface exposes identity, orientation, connection state, containment, and remaining material. A useful instrument interface exposes calibration status and the conditions under which a measurement is valid. A useful action description includes preconditions, resource requirements, expected observations, and permitted responses when these observations disagree. Terrestrial protocol representation and robotic execution provide relevant precedents, while space instruments show the benefits of highly constrained action spaces \citep{grd_steiner2019,pln_sam2012,bio_kidney2024}.

For orbital laboratories, a plausible near-term arrangement combines sealed assay modules, dedicated pumps and actuators, a shared inspection or handling capability, and independent sample-state checks. The shared robot need not manipulate open biological material if exchanging a sealed module supplies the needed flexibility. For planetary fieldwork, the corresponding arrangement combines mobility, target interpretation, instrument placement or sample acquisition, and confirmation of what was actually measured. These designs share a scientific logic but require different mechanics, containment strategies, and environmental validation.

Recovery is a new experimental transition. Recalibration may restore a measurement channel, but a new control cannot reconstruct an unknown past exposure or delivered dose. The controller must distinguish continuation of the original experiment, retention of only eligible endpoints, initiation of a new experiment with a new baseline, and termination. Muscle-chip preservation and MOBP's interrupted operations motivate these distinctions; Curiosity and InSight constrain what mechanical intervention can achieve \citep{bio_muscle2023,bio_mobp2025,rob_curiosity2022,rob_insight2022}.

\subsection{Worked orbital design and a planetary extension}
\label{sec:worked-designs}
The following prospective tests specify candidate sensors, actions, and decision rules for validation under the reported failure and support conditions.

\paragraph{Orbital culture and fluid delivery.}
A concrete endpoint is the flight--ground contrast in \textit{MYH1} transcript abundance, measured after return, in non-electrically stimulated muscle myobundles from a pooled older, sedentary male donor cohort \citep{bio_muscle2023}. This is a transcript endpoint, not a measure of contractile function or proof of a microgravity-specific mechanism. Table~\ref{tab:worked-branch} instantiates a prospective experiment using that endpoint. The experimental unit is one chip; parallel flight and ground chips use the same donor pool, culture age and assigned termination time. Chip replicates do not supply independent donor replication. Temperature and delivered-volume limits, diagnostic calibration and assay-quality criteria must be qualified before testing the proposed controller.

\begin{table}[t]
\centering\small
\caption{A proposed muscle-chip decision branch. The reported study supplies the transcript endpoint, preservation precedent and RNA-quality assessment \citep{bio_muscle2023}; the controller, independent delivery witness and exchange interface are new design requirements for validation.}
\label{tab:worked-branch}
\begin{tabularx}{\linewidth}{P{0.21\linewidth}YY}
\toprule
Step & Complete history: continue & Unresolved delivery: preserve \\
\midrule
Question and observations & Estimate the \textit{MYH1} flight--ground expression contrast at the assigned day-10 termination. Identity, temperature and an independent calibrated delivery record satisfy the predefined culture protocol. & Before day 10, flow telemetry becomes unreliable and a gap in the independent delivery record leaves actual perfusion unknown. Images and command acknowledgements cannot resolve the missing history. \\
Decision and action & Continue culture, then deliver RNAlater and transfer the sealed unit to qualified cold storage at the assigned time; terminate matched ground chips on schedule. & Trigger early preservation through an intact, separately verified reagent path, then transfer to qualified cold storage. If either step cannot be verified, keep preservation status unresolved and seek support. \\
Independent acceptance & An assessor blinded to controller type reconciles sample identity, delivery, thermal and termination records; an independent laboratory checks RNA integrity and sequencing quality before estimating the contrast and its uncertainty. & The same laboratory may accept an RNA measurement at the recorded early termination time if identity, preservation and assay checks pass. Unknown prior perfusion still excludes the planned exposure comparison, even with intact RNA. \\
Retained and lost claims & Retain the specified transcript comparison if all criteria pass; RNA quality alone does not establish valid exposure. & Retain a descriptive molecular endpoint with the interruption recorded. Do not relabel it a day-10 result or infer a flight effect from the unmatched history. Replacement starts a new unit and baseline. \\
\bottomrule
\end{tabularx}
\end{table}

In this design, pumps and a qualified alternate reagent channel perform preservation; a software rule can already select early termination. A shared robot adds value only if the required sealed-unit transfer is outside those mechanisms and timely crew support is unavailable. Compare a dedicated transfer mechanism, rule-based and adaptive policies on the same action set, robotic transfer, and budgeted crew assistance. Every transfer requires validated containment, restraint, connection and storage-state checks. Continuing facility power loss remains outside this action set.

A second branch addresses a scientific question even with healthy hardware. The muscle-chip report identifies the inability to terminate selected samples early while maintaining the others at culture temperature as a limit of its nonmodular design \citep{bio_muscle2023}. Its terminal transcript result therefore motivates a proposed subsequent campaign with separate early and late chip cohorts and time-matched ground controls. Selective preservation and cold transfer would permit comparison of an early transient \textit{MYH1} shift with a sustained or later-emerging shift; independent acceptance follows Table~\ref{tab:worked-branch} at each assigned time. RNA analysis after return informs the next campaign, without assuming an onboard transcript assay. The outcome is the estimated change in the flight--ground contrast between times, with uncertainty; time resolution alone cannot separate launch, microgravity and other exposures causally. Individually addressable preservation modules could supply this scientific action without a shared robot. Robotic transfer is a candidate when separation into different storage conditions is otherwise unavailable.

\paragraph{Planetary acquisition and analysis.}
Consider an investigation relating mineral composition to a mapped geological unit. Its experimental unit includes the target context, acquired material, and the measurement performed. Sample confirmation on Perseverance and the distinct acquisition and analytical functions of SAM motivate separate checks of contact, material retention, and measurement identity \citep{rob_perseverance2021,pln_sam2012}. Proposed observations include pre- and post-action images, calibrated contact telemetry, sample-presence checks, and instrument reference measurements. A completed drilling command cannot establish that a representative sample reached the analyzer; an image of a filled container cannot by itself exclude contamination.

If placement is wrong, repositioning an existing arm may supply the missing action. If material is not retained, a different acquisition procedure may help only within validated tool and contact limits. Curiosity supplies evidence of an alternative procedure enabled by existing mechanics; InSight shows why partial mechanical assistance does not guarantee the intended measurement \citep{rob_curiosity2022,rob_insight2022}. Dust, destination-specific contact forces, thermal exposure along the sample path, and material containment constrain which interventions are feasible. Candidate actions must preserve geological provenance and account for consumed sample and energy.

The original scientific comparison can continue only if identity, context, and analytical conditions remain adequate for the endpoint. Acquiring a new target creates a new unit and may change the sampling distribution; its success must not conceal the abandoned target. A contamination ambiguity may permit contextual imaging while excluding a chemical-abundance claim. Compare fixed acquisition sequences, adaptive policies on the same hardware, a specified added placement or transfer capability, and ground-assisted replanning under explicit communication windows. The hypothesis is that an added action makes a scientifically necessary observation attainable within those windows.

\subsection{Evaluation must connect autonomy to scientific return}
Evaluation should compare complete workflows under the same preassigned scientific tasks and support assumptions. First compare adaptive policies with a predesigned rule controller using the same hardware, observations and permitted actions, including retry, preservation and stopping. Then compare a fixed actuator repertoire with a specified robotic intervention while holding the sensing and policy components constant wherever feasible. If new sensing is inseparable from the robot, an ablation or an explicitly combined-system claim is required; its benefit cannot be attributed to manipulation alone. Include a human-assisted comparator with expert and crew time counted. Report differences that cannot be matched, including installation, calibration, consumables, downtime, energy, mass, and communications. A preserved rare sample may be valuable even without increased throughput.

\begin{table}[t]
\centering\small
\caption{Evaluation priorities for sustained scientific campaigns. Predefine tasks, experimental units, at least one scientific information measure per task, fault cases and resource budgets. Report counts and denominators, including abstentions and failed attempts; thresholds are task-specific and must be justified before comparison.}
\label{tab:metrics}
\begin{tabularx}{\linewidth}{P{0.18\linewidth}YY}
\toprule
Criterion & What to report & Comparison or independent check \\
\midrule
Scientific return and coverage & Task-specific parameter error or interval, hypothesis discrimination, or required geological coverage; valid outcomes and attempted, skipped and abandoned tasks & Preassigned scientific question and budgets; report distinct retained endpoints separately, without treating their counts as equivalent value \\
Scientific acceptance & Valid runs / all attempted runs; false acceptances / all invalid test cases & Endpoint-specific history and measurement criteria; independent or blinded assessment \\
Component benefit & Change attributable to sensing, policy or a specified physical action & Same-hardware policy comparison; matched sensing/policy for action comparison; human comparator \\
Recovery and preservation & Detection, diagnosis, intervention and valid restoration counts; preservation and stops separately & Common fault distribution, including misses; stratify by recoverability relative to declared sensing, actions and support \\
Support burden & Crew/expert effort, consumables, calibration and downtime & Entire campaign, including setup, replenishment and recovery \\
Generalization & Held-out performance across runs, samples, environments and fault combinations & Prevent leakage across related samples and repeated measurements \\
Evidence integrity & Sample lineage, calibration, interruptions, decisions and raw-data links & Independent reconstruction of selected conclusions; retain unresolved history \\
\bottomrule
\end{tabularx}
\end{table}

Benchmarks should include scientific ambiguity as well as mechanical faults. Useful scenarios include a genuine physical instability, a degraded sensor observing a stable process, and a failed actuator producing a misleading measurement. A successful system distinguishes these situations well enough to choose a justified response. Tests should also expose uncertainty that cannot be resolved with the available hardware. Abstaining, preserving material, or escalating can then be correct outcomes. A benchmark that rewards only nominal sequence completion would penalize precisely these scientifically responsible decisions.

Acceptance among attempted runs is insufficient on its own: a controller could improve that fraction by avoiding difficult tasks. Report coverage against the preassigned task set, all exclusions, and support consumed by abstention or escalation. Also report accepted-valid, accepted-invalid, unresolved and rejected endpoints against independent assessments, including valid endpoints incorrectly rejected. Lower false acceptance is useful only with a prespecified requirement for retaining valid evidence; declaring every result unresolved must not count as success.

Estimate the main campaign effect over the same preassigned tasks and fault distribution, including irreversible events, under common resource budgets and support windows. Define recoverability before observing results and relative to an explicit sensing, action and support set: the baseline repertoire, the capabilities shared by both groups, or each group's own repertoire. Rates using different recoverable denominators are explanatory strata, not direct evidence of overall superiority. For irreversible events, report correct detection, preservation or stopping and remaining usable endpoints separately from restoration. Independent checks must assess the claimed endpoint, not merely whether a program resumed.

\begin{figure}[t]
\centering
\includegraphics[width=\linewidth]{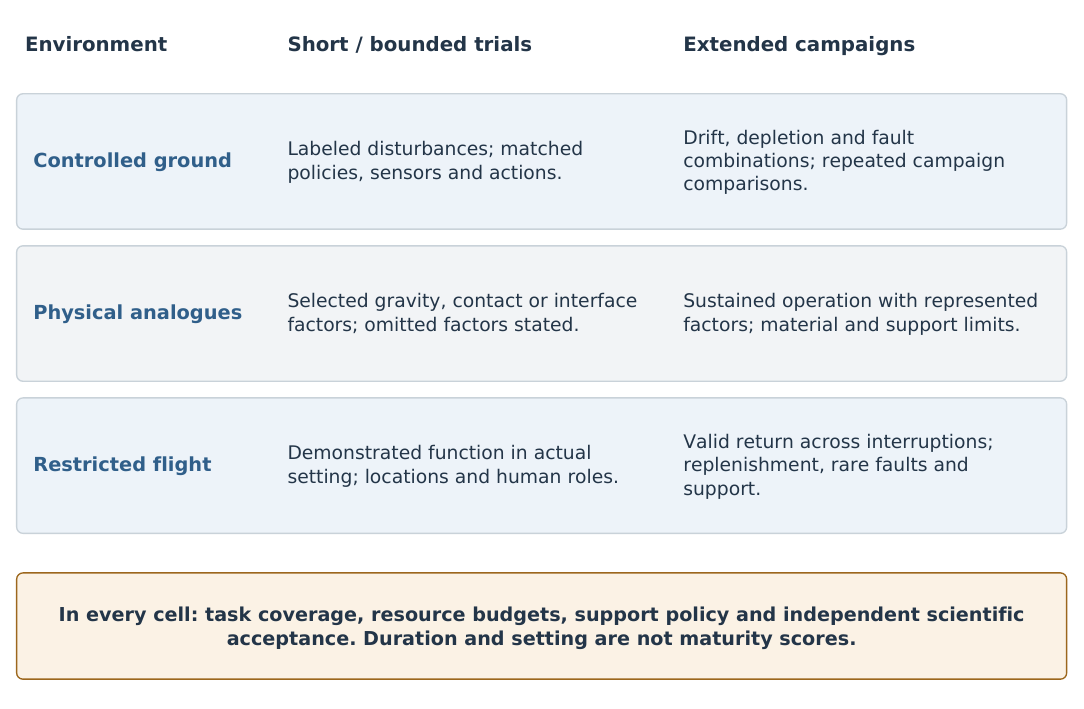}
\caption{Proposed validation matrix. Environmental setting and duration are independent dimensions; extended ground operation and a short flight trial answer different questions. Each cell specifies evidence needed for a prospective validation test, not a readiness score. Task scope, represented environmental factors, and available support must be reported throughout, with dedicated-automation and human-assisted comparators. Original synthesis.}
\label{fig:validation}
\end{figure}

Validation should cross environmental setting with campaign duration and support policy (Fig.~\ref{fig:validation}). Ground tests permit controlled faults and repeated comparisons; analogues test selected physical factors; restricted flight trials test demonstrated functions in their actual operational setting. Extended operation is a separate axis that can be studied in each setting. It exposes drift, contamination, depletion, and combinations of faults missed by a short trial. No setting or duration establishes all the others, and simulation remains separately identified.

\subsection{A research agenda with falsifiable claims}
Three bounded hypotheses follow. First, additional state verification should reduce false scientific acceptance at matched task coverage and budgets while meeting a prespecified valid-endpoint retention requirement; unchanged errors, loss of valid evidence, or excessive diagnostic cost would limit its value. Second, a specified robotic action should improve valid return on tasks whose documented physical bottleneck exceeds the baseline repertoire. Equivalent return from a fixed mechanism, or losses through contamination and resource overhead, would count against that action. Third, diagnosis-linked replanning should retain more eligible endpoints after predefined events than a predesigned rule controller with the same sensing and action set, including retry, preservation and stopping; automatic restart is an additional weak baseline. Superiority to restart alone supports only endpoint-aware recovery, not an advantage from learning or manipulation. This hypothesis fails if gains over the rule controller disappear under independent endpoint assessment or depend on uncounted expert intervention.

These tests locate the design boundary introduced in Section~\ref{sec:intro}. The objective is a system with enough evidence about its experiment and surroundings to continue, adapt, preserve, or stop for scientifically defensible reasons. Space robots become central where their perception and actions expand that feasible set of scientific choices at an acceptable mission cost.

\section{Conclusion}
AI4S in space already encompasses a substantial and heterogeneous body of work. Orbital biology and physical-science experiments provide learning-assisted inference, richer metrology, and capable dedicated automation. Planetary systems demonstrate onboard target selection and adaptive measurement, alongside sophisticated but often human-supervised sampling and recovery. Terrestrial self-driving laboratories show how experimental feedback can accelerate research, while exposing the continuing importance of replenishment, calibration, reliable execution, and independent validation.

Sustained investigation requires a connection between scientific decisions and the physical conditions that make their results interpretable. Controllers need observations of sample and instrument state, executable interventions, and checks that establish which scientific claims survive each transition. Inspection, interface exchange, material repositioning, and preservation can extend a campaign's scope when its existing action repertoire is insufficient.

Space robots become enabling scientific infrastructure when they supply necessary perception and interventions beyond the validated capabilities of an instrument and the timely reach of human support. Their contribution should be evaluated against dedicated mechanisms, redundancy, and human-supported operation through valid scientific return, task coverage, and resource and support costs. Comparisons must retain failed recoveries and unresolved histories. The research agenda is therefore to test which additional physical capabilities expand the scope and continuity of valid science. Connecting that evidence to scientific planning offers a concrete route toward sustained investigation beyond Earth.

\FloatBarrier
\appendix
\section{Search, evidence extraction, and limitations}
\label{sec:methods}
\subsection{Review design and scope}
This is a structured critical narrative review with an annotated evidence map. It asks which scientific functions AI performs in orbital experiments and planetary fieldwork, where data, computation and actions occur, what physical and human support those functions require, and when additional robotic capability could improve valid scientific return. Selection is purposive, rather than exhaustive. The review does not estimate field-wide prevalence, pool effects, or assign a formal risk-of-bias score. Astronomy and remote sensing provide historical contrasts; generic spacecraft autonomy is included only where it clarifies a scientific workflow or its operational support.

Searches were expanded on 25 October 2026 after an initial scoping pass on 24 October. Publication eligibility was bounded at 25 October 2026; targeted source, correction and methodological checks continued during revision on 26 October. Historical planning and experimental-automation work was eligible. The PubMed queries used a technical date bound of 1 January 1900 to the cutoff, rather than a recent-publication restriction. Journal versions were preferred for citation when available, while the actually inspected version---including author preprints---was recorded. Proposed, simulated, ground, analogue and flight work remain distinguishable.

\subsection{Search channels and selection rationale}
Search channels included PubMed, targeted web discovery, publisher and author pages, arXiv, NASA Technical Reports Server, and NASA, ESA, JAXA, CNSA and other institutional documentation. Crossref and Europe PMC supplied targeted bibliographic and access checks. This was not a complete search of every named service; no complete Scopus or Web of Science export or comprehensive multilingual search was performed. Search families paired scientific settings with analytical or experimental functions: space biology with learning, omics, imaging, culture and microfluidics; physical experiments with learning or optimization; planetary science with target selection, adaptive spectroscopy, sampling and scientific planning; and terrestrial laboratories with experimental decisions, robotic execution, validity, recovery and human intervention. Named-system queries and relevant references supplied additional sources.

Four PubMed query families returned 344 query hits and 328 unique identifiers: space-biology AI (111), orbital biological automation (83), biological nanosatellite heritage (94), and space-biology data resources (56). Thirty records entered the focused biology synthesis. These are channel-specific retrieval and selection counts, not eligible-study or full-text-exclusion counts. The screening table records title screening and abstract assessment for selected candidates. The 298 unselected records have generic notes rather than reassessed, record-specific exclusion reasons.

Selection sought contrastive scientific functions and support conditions: direct AI contributions, dedicated-automation comparators, physical bottlenecks, documented interventions or failures, terrestrial transfer evidence, and necessary context. Examples such as bounded onboard targeting, fixed assay devices, ground-designed recovery, failed penetration and incomplete experimental histories test different parts of the argument. They are not matched comparative trials. Multiple reports of one platform are linked rather than counted as independent demonstrations. Search expansion was bounded by these questions, the source cutoff and the preparation period; no prespecified saturation test, database-wide screening endpoint or citation quota was used. The supplementary selection rationale is retrospective where the original selection sequence was not recorded.

\subsection{Evidence roles, coding and source access}
The evidence set contains \ReviewRecordCount{} bibliographic records: \ReviewCoreAICount{} coded as direct scientific AI evidence, \ReviewSupportingCount{} as supporting automation or operational comparators, \ReviewTransferCount{} as terrestrial transfer evidence, and \ReviewBackgroundCount{} as background. These roles describe how each record is used in the synthesis; study type, deployment setting and autonomy are separate attributes. Direct scientific AI evidence includes ground analysis of space-derived data, engineering tests, simulations and field analogues. Counts refer to bibliographic records rather than independent flight deployments, and include reviews, mission accounts, a linked correction and procedural provenance. Related reports retain shared system identifiers, with experimental and analytical phases distinguished.

The codebook separates data origin, computation location, action location, physical operations and human roles. Post-acquisition analysis does not by itself establish a geographical location. A location inferred from a workflow is labeled as inferred; unknown details remain unknown. Not applicable refers to a question outside the record-level claim, such as a new physical intervention in a software-only analysis, and does not mean that the wider platform lacks that capability. The evidence supplement retains the raw extraction and field-level change records, including separate coding of biological operations and human support. Some legacy source locations and intervention details remain unresolved.

Figure~\ref{fig:evidence-map} applies the codebook to selected workflows. Records sharing a system identifier remain distinct by experimental or analytical phase unless an explicit mapping links them; these mappings are documented in the evidence supplement.

Original access descriptions are retained alongside coarse navigation categories. Explicit missingness codes distinguish unknown, partly known and inapplicable fields; the source description remains authoritative for their interpretation. An abstract can support a bounded statement; downloaded but unassessed full text is not counted as a full-text review. Primary studies support scientific-performance claims, while agency accounts can establish operational events. Selected passages were rechecked for deployment, metric definitions, human support and corrections; the A-Lab synthesis uses its corrected publication, and the CAL axis correction is recorded. These checks are neither independent human dual screening nor replication or exhaustive correction surveillance.

\subsection{Supplementary audit materials and limitations}
The accompanying ancillary evidence files contain the raw and revised matrices, codebook, selection rationale, query and access logs, original screening decisions with revision annotations, source-check updates, and normalization script. Record-to-log links distinguish explicit query links from identifier matches in retrieval or verification logs; the latter do not establish which query first discovered a source. Initial discovery paths that cannot be reconstructed remain marked unresolved.

The K-Dense \texttt{literature-review} skill was a procedural aid, with its associated resource paper cited for provenance \citep{bg_kassis2026}. Generative AI assisted search, drafting, coding and cross-checking; this assistance does not constitute independent scientific validation. Coverage is limited by accessible-source bias, reporting gaps, uneven agency documentation and the emphasis on support and recovery. Unreported interventions leave support requirements unresolved, and the proposed architecture and evaluation agenda require prospective testing.

\begingroup
\raggedright
\bibliographystyle{unsrtnat}
\bibliography{references}
\endgroup
\end{document}